\documentclass[letterpaper, 10 pt, conference]{ieeeconf}

\usepackage[T1]{fontenc}
\usepackage[utf8]{inputenc}

\IEEEoverridecommandlockouts

\usepackage{graphicx}
\usepackage{capt-of}
\usepackage{amsmath}
\usepackage{amssymb}
\usepackage[table,dvipsnames]{xcolor}
\usepackage{booktabs}
\usepackage{makecell}
\definecolor{metricorange}{HTML}{F4A261}

\newcommand{\robotname}{Pixel}

\title{\LARGE \bf
  Pixel: A Co-Designed Full-Stack Robot System for \\ Learning General-Purpose Mobile Manipulation
}

\author{
  Tony Tao\textsuperscript{*},
  Jason Jingzhou Liu\textsuperscript{*},
  Ruslan Salakhutdinov$^{\dagger}$,
  Deepak Pathak$^{\dagger}$\\[0.5em]
  Carnegie Mellon University\\
  \textsuperscript{*}Equal contribution \qquad $^{\dagger}$Equal advising\\[0.4em]
  \href{https://pixel-robot.com}{\textbf{\texttt{https://pixel-robot.com}}}
}

\IEEEaftertitletext{%
  \vspace{-6pt}
  \begin{minipage}{\textwidth}
    \centering
    \includegraphics[
      width=\linewidth,
      keepaspectratio
    ]{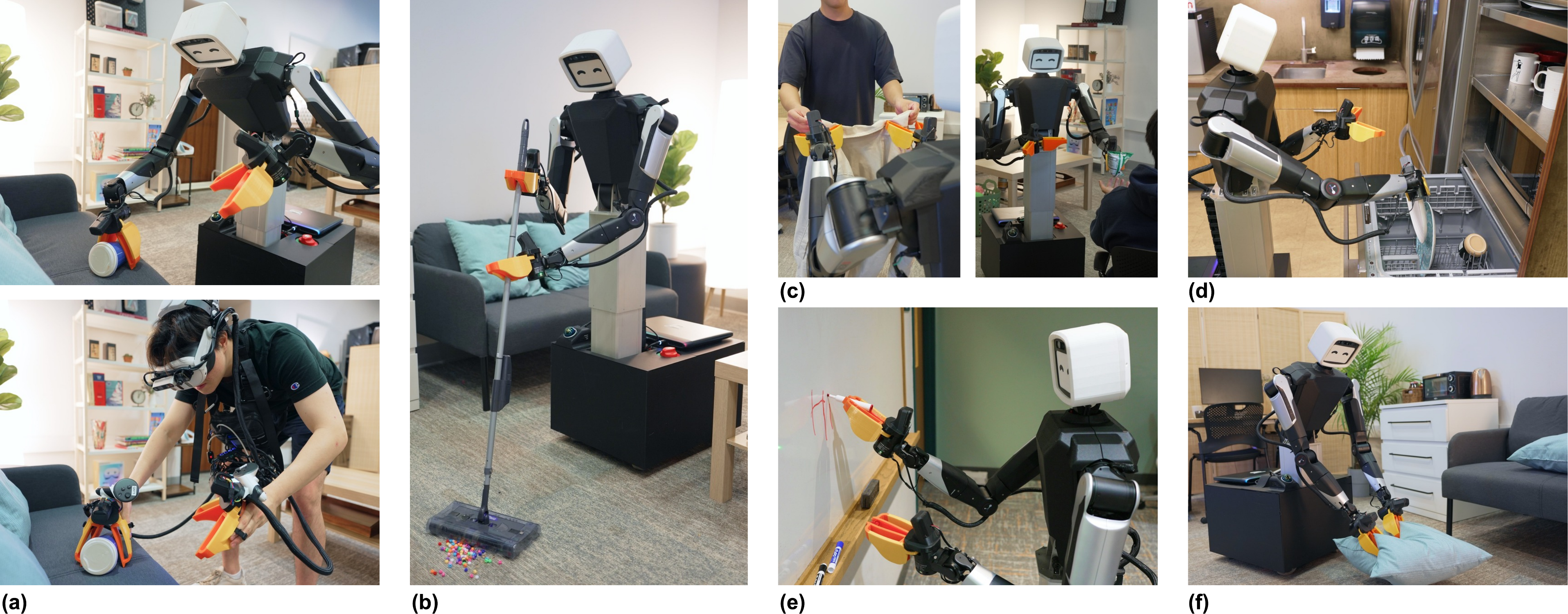}
    \captionof{figure}{\textbf{\robotname{} system highlights.} (a) Our co-designed data-collection system captures human whole-body demonstrations and transfers them to our custom robot platform. (b) A whole-body controller coordinates motion across all robot joints. (c) Natural human--robot interaction. (d) Operates seamlessly in human environments. (e) Dexterous, low-latency, compliant teleoperation. (f) Human-like functional kinematics provides extended workspace.}
    \label{fig:teaser}
  \end{minipage}
  \vspace{0.3em}
}

\makeatletter
\let\NAT@parse\undefined
\makeatother

\usepackage[
  colorlinks=true,
  linkcolor=BurntOrange,
  citecolor=BurntOrange,
  urlcolor=Orange
]{hyperref}

\begin{document}
\maketitle
\thispagestyle{empty}
\pagestyle{empty}

\begin{abstract}

  General-purpose mobile manipulation requires a combination of capable robot hardware, scalable demonstration collection, expressive whole-body control, and effective policy learning. Yet these components are often developed independently, creating mismatches that limit the capabilities of the complete system. We present \textbf{\robotname{}}, a co-designed full-stack system for learning general-purpose mobile manipulation. \robotname{} combines a low-cost wheeled humanoid with human-aligned kinematics and compliant grippers, multiple teleoperation and robot-free human data-collection interfaces, a unified whole-body controller, and a common policy learning pipeline across human and robot demonstrations. A hierarchical learning stack further enables language-conditioned visuomotor control and long-horizon task execution. We evaluate the system across hardware, data collection, and policy learning, and demonstrate mobile-manipulation behaviors requiring dexterity, active perception, navigation, and whole-body coordination. We hope that \robotname{} provides an integrated and extensible foundation for studying how robot hardware, data, control, and learning can be designed together for robot learning. More results at \href{https://pixel-robot.com/pixel/}{\texttt{https://pixel-robot.com/pixel}}

\end{abstract}

\section{INTRODUCTION}
\label{sec:introduction}

Imagine asking a robot to tidy your bedroom. It might need to collect clothes from the floor, place objects into cabinets at different heights, or operate a vacuum. The robot might also need to understand your instructions, determine where objects belong, plan a sequence of behaviors, and adapt when you rearrange the furniture.

Realizing this vision requires integration across a deep robot-learning stack. The hardware must provide the mobility, workspace, strength, compliance, and sensing needed to interact safely in human environments. Its morphology determines where it can reach and how effectively it can leverage different data sources, while its end effectors determine which objects and tools it can operate. Broad capability also requires large, diverse datasets and the interfaces that make such data practical to collect. Yet demonstrations are useful only if the robot's controllers can translate them into safe, physically realizable motions and learning algorithms can extract behaviors that generalize to new situations. Maximizing system-level capability therefore requires co-design across these layers, such that no individual component becomes the bottleneck on the performance of the complete system.

Thus far, most robotics work focuses on individual parts of this stack. Open-source hardware platforms have made capable robots and end effectors increasingly accessible, but commonly leave scalable data collection, policy training, and deployment to downstream integration~\cite{shaw2023leap,zorin2025ruka,anjaria2026yor}. Conversely, sensorized human interfaces enable demonstrations to be collected at scale without a physical robot, but transfer to a target platform remains constrained by its morphology, sensing, end effectors, and controllers~\cite{chi2024universal,zeng2025activeumi,TaoT-RSS-25}. Other systems connect hardware, teleoperation, and policy training to achieve impressive learned manipulation~\cite{zhao2023learning,fu2024mobile}, but use comparatively simple embodiments and rely primarily on demonstrations collected through teleoperation. What remains underexplored is a mobile manipulation system in which the embodiment, data interface, controller, action representation, and learning stack are designed jointly from the start.

\begin{figure*}[t]
  \centering
  \includegraphics[
    width=\textwidth,
    keepaspectratio
  ]{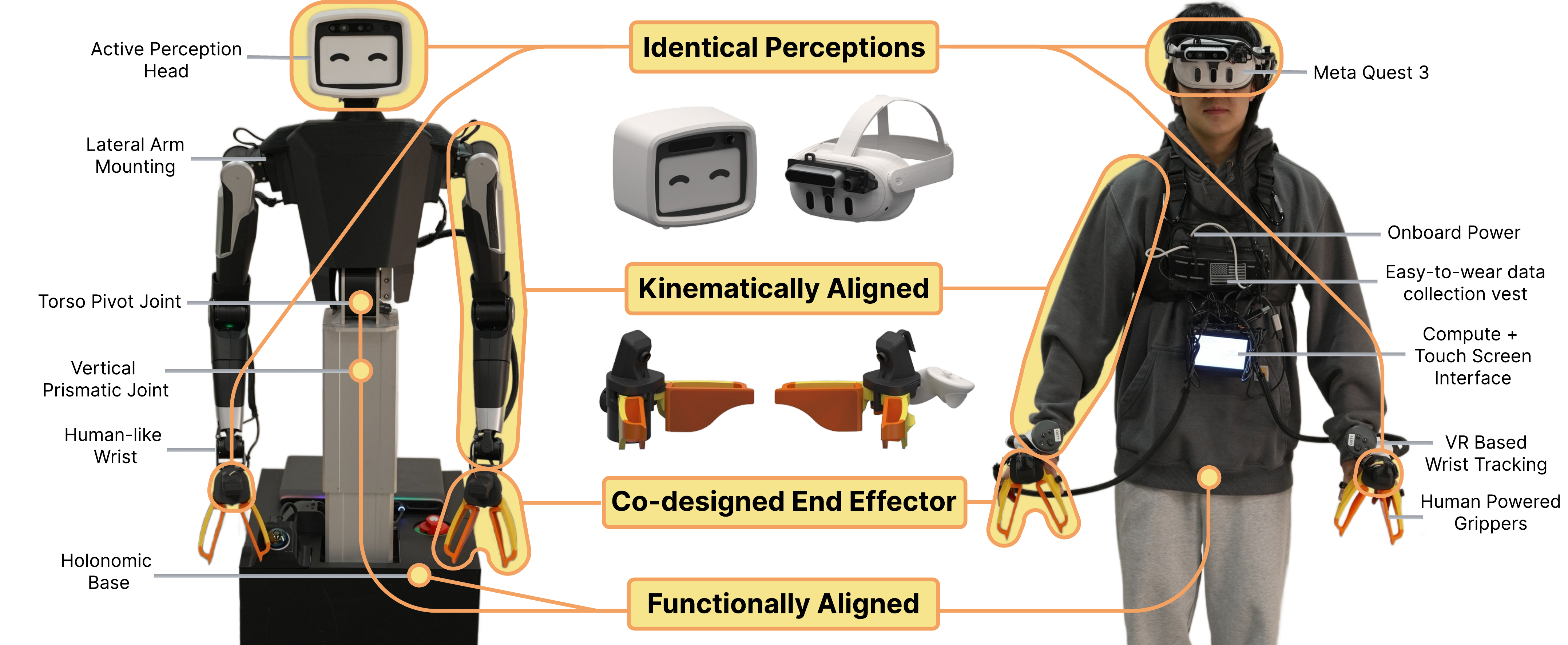}
  \caption{\textbf{Hardware co-design.} Our robot hardware and sensorized human data collection system are co-designed at all levels, with shared camera perspectives, aligned human--robot kinematics, paired end effectors designed from the ground up, and a functionally aligned lower-body design.}
  \label{fig:hardware_codesign}
  \vspace{-4mm}
\end{figure*}

In this work, we present \textbf{\robotname}, a co-designed, open-source mobile robot and learning stack that enables broadly capable mobile manipulation. At the center of the system is a wheeled humanoid robot costing under \$15{,}000. Its human-proportioned 17-DoF upper body is designed to align with human motion, while a holonomic mobile base, vertical lift, and articulating torso provide wide workspace coverage. Wide-aperture, high-force robot grippers are paired with lightweight, comfortable sensorized human-worn grippers, allowing demonstrations to be collected in the wild while retaining high action fidelity.

This co-design extends from the hardware to the data-collection interfaces, control, and learning. Leader-arm teleoperation, VR teleoperation, and sensorized human demonstrations all produce task-space targets that are mapped to the robot through a shared whole-body controller that coordinates motion while accounting for joint limits and self-collision avoidance. We also introduce a common relative-base action space which represents motion consistently across human and robot demonstrations, enabling flexible policy training across data sources.

We systematically ablate design choices across hardware, data-collection interfaces, and model training. Policies trained from sensorized human demonstrations achieve an average task progress of 82.0\% across three challenging mobile-manipulation tasks, while policies trained from robot-teleoperated demonstrations achieve 78.7\% across three additional tasks.

By co-designing the robot and learning stack, \robotname{} supports rapid iteration across embodiment, data, learning, and deployment within a single extensible platform.

\section{RELATED WORK}
\label{sec:related_work}

\subsection{Open-Source Robot Platforms}
\label{sec:related_robots}

Open-source robot platforms accelerate progress by lowering entry costs and enabling researchers to repair, extend, and iterate on others' designs. Yet most prior work addresses only one layer of the robot-learning stack, whether hardware components such as end effectors~\cite{shaw2023leap,zorin2025ruka, ma2017yale, bhirangi2023all, christoph2025orca}, demonstration interfaces for data collection~\cite{wu2024gello, chi2024universal, fang2024airexo, ohkawa2026yubi, liu2025factr,oh2026factr2}, or data and training infrastructure~\cite{cadene2026lerobot,allshire2026abc, mandlekar2022matters, openx2024embodiment, kim2025openvla}. These projects broaden access, but building a complete system still requires integrating independently designed hardware, sensing, control, data collection, and learning infrastructure, often developed with different objectives and assumptions.

Several platforms integrate multiple layers of the stack but retain important limitations. Some platforms offer more complete hardware systems but lack the morphology required for general-purpose mobile manipulation~\cite{wu2024tidybotopensourceholonomicmobile, kemp2022stretch, chandaka2025lekiwi, wang2025xlerobot}. Other open platforms demonstrate impressive physical capabilities but either lack policy-training and deployment infrastructure~\cite{anjaria2026yor} or emphasize locomotion over capable manipulation~\cite{liao2025berkeley,chi2025berkeleylite}. Systems that combine hardware, data collection, and policy training often rely heavily on teleoperation, limiting the scalability of data collection~\cite{zhao2023learning,fu2024mobile, shi2025toddlerbot,robotics2026fauna}.

\robotname{} aims to provide a complete, scalable system by co-designing all parts as one extensible stack.

\subsection{UMI Systems \& Human Data Collection}
\label{sec:related_umi}

Human data-collection interfaces have grown rapidly in popularity because they enable data to be gathered easily and at scale without a robot. As a more scalable alternative to robot teleoperation, they have seen widespread adoption in both research and industry~\cite{sunday2025act1,generalist2025gen0}.

Sensorized human data collection, however, introduces an embodiment gap: human actions must map to executable robot actions. Robot-compatible tools minimize this gap through a matching interface, ranging from early gripper-in-hand devices~\cite{song2020grasping,pmlr-v155-young21a,shafiullah2023bringing} to the UMI family~\cite{chi2024universal,pmlr-v305-zhaxizhuoma25a,zeng2025activeumi,ohkawa2026yubi}. Other methods trade direct alignment for more natural collection: DexUMI constrains the hand with an exoskeleton~\cite{xu2025dexumi}; DexCap and DexWild use wearable gloves with retargeting~\cite{Wang-RSS-24,TaoT-RSS-25}; and EgoMimic captures bare-hand egocentric video~\cite{kareer2025egomimic}. These designs balance natural interaction against action fidelity.

While most prior systems capture primarily hand motion and target tabletop manipulation, recent work has begun using human demonstrations to learn richer behaviors involving active perception, whole-body coordination, and mobile manipulation. ActiveUMI and EgoMI jointly capture head and hand motion to enable learning of active-perception behaviors~\cite{zeng2025activeumi,yu2025egomi}. For mobile manipulation, EMMA co-trains on egocentric full-body human demonstrations and static robot data, while HoMMI learns bimanual navigation and manipulation from head-and-gripper UMI trajectories~\cite{zhu2026emma,xu2026hommi}. HuMI and HALOMI further extend human data collection to commercial humanoids by incorporating lower-body motion~\cite{nai2026humanoid,zhao2026halomi}. However, the co-design of these systems is generally localized to selected components or implemented on an existing robot embodiment.

\robotname{} jointly optimizes the complete system for seamless transfer from human demonstrations to whole-body robot skills.

\subsection{Behavior Cloning for Mobile Manipulation}
\label{sec:related_mobile_bc}

Behavior cloning has become a widely used approach to robot policy learning, driven in part by advances in policy architectures and action representations. ACT introduced action chunking to improve temporal modeling~\cite{zhao2023learning}; Diffusion Policy brought diffusion models to robotics, enabling policies to represent multimodal action distributions~\cite{Chi-RSS-23}; and UMI introduced relative end-effector actions to improve learning when no consistent absolute reference frame is available~\cite{chi2024universal}.

Compared with tabletop manipulation, mobile manipulation can introduce additional challenges for behavior cloning, including tighter coupling between perception and action, compounding errors over longer horizons, and the difficulty of deploying learned policies on complex robotic systems. In the simplest setting, when demonstrations directly contain robot actions, policies can learn coordinated whole-body motion end to end~\cite{fu2024mobile,wu2024tidybotopensourceholonomicmobile,BlackK-RSS-25}. However, this approach requires collecting demonstrations on the complete robot and assumes that teleoperating the robot through the desired motions is feasible, which can itself be challenging.

Direct imitation of robot actions, however, is not always feasible. On legged robots, small joint-prediction errors can violate balance or contact constraints, while human demonstrations do not provide robot joint states. Many systems therefore predict task-space motion and delegate joint coordination to retargeting or whole-body control. Legged platforms pass learned end-effector trajectories to controllers that maintain balance~\cite{he2024learningvisualquadrupedallocomanipulation,ha2024umionlegs}, while wheeled and humanoid systems map keyposes or human head, hand, pelvis, and foot motion to feasible whole-body actions~\cite{sundaresan2025homer,zhu2026emma,xu2026hommi,nai2026humanoid,zhao2026halomi}.

Long-horizon tasks may also require structure beyond low-level action prediction. Hi Robot decomposes instructions into atomic language commands executed by a low-level policy~\cite{pmlr-v267-shi25d}; CoT-VLA predicts visual subgoals before generating action chunks~\cite{Zhao_2025_CVPR}; and $\pi_{0.7}$ uses language, strategy metadata, and visual subgoals to steer and compose behaviors across tasks~\cite{intelligence2026pi07steerablegeneralistrobotic}.

\section{SYSTEM DESIGN}
\label{sec:system_design}

We present \textbf{\robotname{}}, a low-cost wheeled humanoid and integrated
robot-learning stack for general-purpose mobile manipulation. Rather
than developing the robot embodiment, data-collection interfaces, and
learning infrastructure independently, we design them jointly so that
each component complements the others and the robot can realize its
full capability as one complete system. Figure~\ref{fig:hardware_codesign} summarizes the hardware co-design.

In general, our design is guided by three principles:
\begin{enumerate}
    \renewcommand{\theenumi}{G\arabic{enumi}}
    \renewcommand{\labelenumi}{\textbf{\theenumi.}}

  \item \label{goal:hardware}
    \textbf{Cost-effective, reproducible, and safe.}
    The system should maximize manipulation capability while minimizing hardware cost and complexity, remain easy to reproduce, maintain, and extend, and be safe to operate in human environments.

  \item \label{goal:alignment}
    \textbf{Scalable and high-fidelity data collection.}
    The robot and data-collection interfaces should support scalable, high-fidelity data collection and facilitate learning from multiple data sources, including teleoperated and human demonstrations.

  \item \label{goal:capability}
    \textbf{Broad everyday mobile-manipulation capability.}
    The robot embodiment and learning stack should support a broad
    range of everyday mobile manipulation behaviors, including dexterous, long-horizon, and whole-body
    tasks.
\end{enumerate}

In the following sections, we describe the
three co-designed layers of the system: the robot hardware, the
data-collection interfaces, and the policy-learning stack.

\subsection{Robot Hardware System}
\label{sec:hardware}

\begin{figure*}[t]
  \centering
  \includegraphics[
    width=\textwidth,
    keepaspectratio
  ]{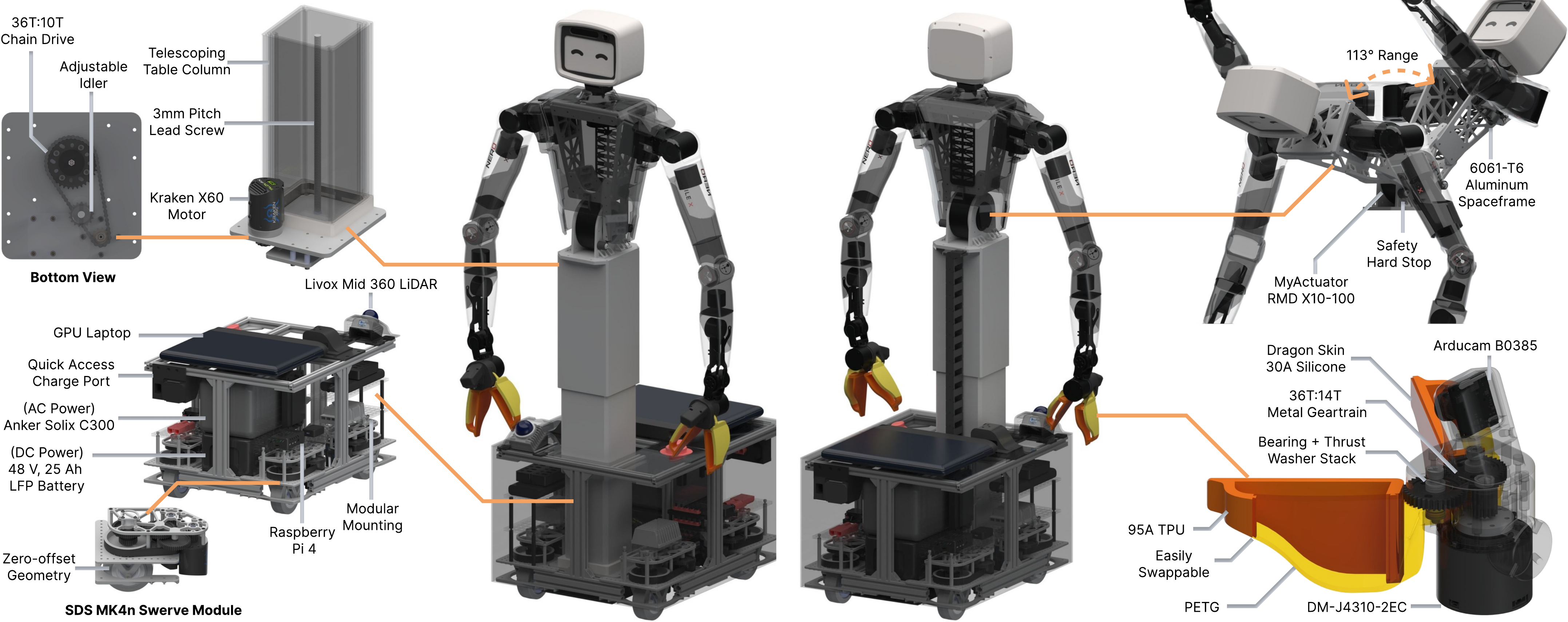}
  \caption{\textbf{\robotname{} hardware overview.} Our low-cost robot platform is designed for high performance, reliability, and reproducibility. The system integrates a holonomic mobile base, telescoping vertical lift, articulated torso, active-perception head, and custom high-force, compliant end effectors.}
  \label{fig:hardware_details}
  \vspace{-4mm}
\end{figure*}

We design \robotname{} to support a wide range of everyday mobile-manipulation tasks in the open world, while making deliberate tradeoffs to balance our design goals \ref{goal:hardware}--\ref{goal:capability}. At the architectural level, we choose to retain the task-relevant advantages of a humanoid morphology while minimizing the mechanical and control complexity of a full humanoid robot. A human-aligned upper body provides a human-like bimanual workspace for interacting with objects, tools, and environments designed for people, while also facilitating the transfer of human demonstrations in support of \ref{goal:alignment}. At the same time, a humanoid system with bipedal locomotion and multi-fingered hands would introduce substantial cost, control complexity, and maintenance concerns. Consistent with \ref{goal:hardware}, we therefore replace legs with a holonomic mobile base and passively stable vertical lift, and use simple custom grippers in place of dexterous hands. These choices trade some terrain versatility, lower-body articulation, and in-hand dexterity for simplicity, while preserving most of the mobility, vertical reach, and grasping capability needed to support the diverse whole-body and dexterous behaviors targeted by \ref{goal:capability}. \textit{The complete hardware platform, shown in Figure~\ref{fig:hardware_details},
can be assembled for under \$15,000.}

\textbf{Holonomic Mobile Base.} \robotname{} is built on a custom holonomic mobile base. Our base has a footprint of $45.5\,\mathrm{cm}\times53.5\,\mathrm{cm}$, compared with $50.0\,\mathrm{cm}\times54.0\,\mathrm{cm}$ for TidyBot++~\cite{wu2024tidybotopensourceholonomicmobile}. This compact footprint allows the robot to navigate cluttered indoor environments and pass through standard doorways. Holonomic mobility enables independent planar translation and rotation, improving maneuverability in confined indoor environments~\cite{wu2024tidybotopensourceholonomicmobile}, providing a simple action space for whole-body teleoperation and policy learning, and facilitating the transfer of natural human whole-body motion to the robot in support of \ref{goal:alignment}.

The chassis is constructed from aluminum extrusions, forming a rigid
superstructure that houses the batteries, power electronics, drivetrain,
lift actuator, and networking hardware. The drivetrain consists of four
MK4n swerve modules with Colson wheels from Swerve Drive
Specialties~\cite{swerve_drive_specialties}. Motivated by our goal of reproducibility in \ref{goal:hardware}, we exclusively use off-the-shelf parts.

Unlike the offset powered-caster design used by TidyBot++, we use zero-offset swerve modules, whose steering axes approximately intersect the wheel contact patches, thus decoupling the steering from chassis motion. This decoupling of steering and chassis translation is particularly important for mobile manipulation, where small steering inputs to the base can cause jerky motions in the end-effector pose and compromise more precise tasks. The resulting drivetrain does not require additional CNC-machined parts \cite{wu2024tidybotopensourceholonomicmobile} and provides simpler kinematic control. Base control details are available in Appendix~\ref{sec:appendix_base}.

\textbf{Vertical Lift.} A large vertical workspace is essential for operating
in human environments, where tasks may require reaching from the floor to high
shelves. An anatomically accurate, multi-joint lower body would more closely
match human kinematics and therefore better satisfy
\ref{goal:alignment}, but would substantially increase the robot's mechanical
and control complexity and cost. Given the mobility of the base, such additional
articulation would provide only limited additional task-relevant workspace
compared with a simpler vertical degree of freedom as shown in Figure~\ref{fig:workspace_compare}. We therefore design a
telescopic lift that provides $430\,\mathrm{mm}$ of travel, a maximum speed of
$75\,\mathrm{mm/s}$, and a rated payload of $40\,\mathrm{kg}$. This design
preserves the task-relevant vertical motion needed to transfer human
demonstrations while substantially expanding the workspace of both arms with a
simple and stable mechanism, balancing
\ref{goal:alignment} and \ref{goal:hardware}.

The lift repurposes a commercial electric table column with a custom drivetrain.
A Kraken X60 motor drives the column's lead screw through a chain transmission,
using a 10-tooth motor sprocket and a 36-tooth screw sprocket for a $3.6{:}1$
reduction. Consistent with the safety objective in
\ref{goal:hardware}, the lead-screw mechanism passively holds its position when
unpowered, preventing the lift and its payload from falling in the event of
power loss. Represented as a single prismatic degree of freedom, the lift also
integrates naturally into the whole-body controller described in
Section~\ref{sec:control_interfaces}. Lift sizing and control details are provided in Appendix~\ref{sec:appendix_lift}.

\textbf{Torso, Head, and Arms.} To facilitate the transfer of human demonstrations to robot motion, \robotname{} uses a human-proportioned, 17-DoF upper body comprising a torso-pitch joint, two 7-DoF arms, and a 2-DoF head. Compared with mobile manipulators whose arms and cameras are rigidly mounted to the lift structure~\cite{anjaria2026yor,fu2024mobile, i2rt_linearbot}, these additional degrees of freedom allow the robot to easily reposition both its manipulation workspace and sensing orientation. In particular, torso articulation extends the forward and downward reach of both arms, while the actuated head enables the robot to maintain task-relevant regions within view during manipulation. These degrees of freedom expand the robot's whole-body manipulation capability (\ref{goal:capability}) while allowing it to reproduce torso pitching and gaze behaviors present in human demonstrations (\ref{goal:alignment}).

To avoid unnecessary mechanical and manufacturing complexity, the upper body is built around a simple, rigid torso structure consistent with \ref{goal:hardware}. The torso is assembled from waterjet-cut, $6.35\,\mathrm{mm}$-thick 6061-T6 aluminum plates and is cantilevered from a single torso-pitch joint driven by a MyActuator X10-100 actuator with a rated torque of $50\,\mathrm{N\,m}$. Coordinating this joint with the vertical lift and arms allows the robot to reach low or recessed objects while maintaining a stable base configuration. To reduce the kinematic mismatch between human and robot motion (\ref{goal:alignment}), we use two AgileX NERO arms, each with 7 DoFs, a rated payload of \(3\,\mathrm{kg}\), and human-like kinematics. The redundant arm degree of freedom provides additional flexibility for satisfying end-effector targets while avoiding joint limits, self-collisions, and unfavorable configurations in confined bimanual workspaces. The head is actuated in yaw and pitch by two DAMIAO DM-J4310-2EC actuators, allowing the onboard cameras to be oriented independently of the torso.

The torso, head, and arms are controlled using joint-space impedance control. For each joint $j$, the commanded torque is
\[
  \tau_j =
  k_{p,j}\left(q_j^\star-q_j\right)
  +
  k_{d,j}\left(\dot{q}_j^\star-\dot{q}_j\right)
  +
  \tau_{g,j}(\mathbf{q})
  +
  \tau_j^{\mathrm{ff}},
\]
where $q_j^\star$ and $\dot{q}_j^\star$ are the desired joint position and velocity, $k_{p,j}$ and $k_{d,j}$ are the corresponding stiffness and damping gains, $\tau_{g,j}(\mathbf{q})$ compensates for gravitational torque, and $\tau_j^{\mathrm{ff}}$ denotes any additional feedforward torque.

\textbf{End Effector.} Everyday manipulation requires both precise handling of small objects and secure grasping of large or heavy objects. Although multi-fingered hands provide additional dexterity, they introduce considerable mechanical and control complexity. At the same time, the end effector should enable intuitive and scalable data collection. We therefore co-design a simple, single-DoF pincer gripper and a corresponding human-worn demonstration interface, balancing the simplicity and reproducibility of \ref{goal:hardware} with the high-fidelity data collection required by \ref{goal:alignment}. The gripper follows a finger-aligned design resembling the natural pinch motion of the thumb and index finger, making it intuitive for human demonstration while remaining capable of grasping a diverse range of objects during robot deployment~\cite{ohkawa2026yubi}. The human-worn interface preserves the task-facing finger geometry and wrist-camera placement of the robot gripper, reducing observation and action mismatch between demonstration and deployment~\cite{chi2024universal}. We describe the wearable interface in further detail in Section~\ref{sec:control_interfaces}.

On the robot side, each gripper is driven by a DAMIAO DM-J4310-2EC actuator through a 14-tooth pinion and two 36-tooth finger gears, yielding a \(2.57{:}1\) reduction. At a motor torque of \(3\,\mathrm{N\,m}\), the transmission provides a nominal combined torque of \(7.71\,\mathrm{N\,m}\) across the finger pivots and a measured closing force exceeding \(90\,\mathrm{N}\). The pincer mechanism provides a maximum aperture of \(240\,\mathrm{mm}\) while maintaining a compact and mechanically simple construction.

Each finger consists of a rigid 3D-printed PETG structure, a compliant 95A-TPU membrane, and a Shore-A 30 Dragon Skin silicone contact surface. The distal portion is narrow and comparatively rigid, enabling precise grasps of small and thin objects. The wider, compliant proximal region conforms to larger objects and handles, increasing contact area and resistance to slip. Separating the rigid load-bearing structure from the compliant membrane allows the finger geometry and surface compliance to be tuned independently. Because deformation of the compliant membrane is visible in the wrist-camera
image, it also provides visual cues about contact state and grasp force without requiring dedicated tactile sensors, providing additional information to the policy in support of \ref{goal:capability}.

The policy commands the gripper using continuous aperture targets, which are tracked using an impedance controller with variable stiffness, damping, and torque limits. Lower gains provide compliant contact for delicate objects, while higher gains enable secure power grasps. Each robot-mounted gripper weighs \(680\,\mathrm{g}\) and can be assembled for under \$200.

\textbf{Sensing, Compute, and Electrical Architecture.} Mobile manipulation in the open world requires tightly integrated sensing, computation, communication, and power delivery. \robotname{} therefore uses a fully onboard architecture designed to support real-time control and perception (\ref{goal:capability}) while remaining modular, low-cost, and easy to maintain
(\ref{goal:hardware}).

The robot's sensing suite consists of three Arducam B0385 global-shutter RGB cameras (one on each wrist and one on the head), a head-mounted Intel RealSense D435 RGB-D camera, and a base-mounted Livox Mid-360 LiDAR. The RGB cameras stream camera-native MJPEG at $640 \times 480$ and 100 FPS, with a wide field of view of approximately $160^{\circ}$ diagonally. Global shutters reduce motion artifacts, and the onboard JPEG compression lowers the processing and storage overhead of multi-view data collection. The wrist cameras capture close-range contact information, and the head camera supplies broader scene context. The D435 provides aligned RGB and depth, which enables 3D reconstruction of the scene and the training of policies based on depth images or point clouds. The LiDAR and its integrated IMU can be used for SLAM or navigation (\ref{goal:capability}).

Computation is divided between a Raspberry Pi 4 and an onboard GPU laptop. The Raspberry Pi controls the drivetrain and lift, whereas the laptop handles whole-body control, data collection, and policy inference. The two computers communicate using ROS~2 over a private wired Ethernet network, with a Wi-Fi access point supporting wireless VR and leader-arm teleoperation. Motor communication is divided across four CAN networks: one for the drivetrain and lift, one for the torso and head, and one for each arm and gripper.

To accommodate devices with different power requirements, \robotname{} uses two complementary onboard power sources. A $48\,\mathrm{V}$, $25\,\mathrm{Ah}$ lithium iron phosphate (LFP) battery supplies the robot's DC electronics, while a $288\,\mathrm{Wh}$ portable AC power source rated at $300\,\mathrm{W}$ powers the GPU laptop, networking equipment, and other peripherals through their standard power adapters. This separation allows compute, sensing, and networking hardware to be added or replaced without redesigning the robot's DC power distribution, supporting the extensibility objectives of \ref{goal:hardware}. LFP was chosen for its long cycle life and favorable thermal stability during frequent operation, making it a safe and reliable choice consistent with \ref{goal:hardware}. Together, the two sources provide approximately $2.5$--$3$ hours of untethered operation, with runtime primarily limited by the smaller AC source and the GPU laptop's power consumption.

\subsection{Whole-Body Control and Data Collection Interfaces}
\label{sec:control_interfaces}

Everyday tasks such as retrieving objects from the floor, loading a dishwasher, and opening a drawer require coordinated motion of the whole body. Supporting these behaviors on a $23$-DoF platform presents challenges for both control and data collection. For many tasks, teleoperators must command the entire robot to produce coherent whole-body motions. Sensorized human demonstrations also need to be mapped effectively to robot motions without losing task-relevant behavior.

Following our goal to facilitate learning from multiple sources in \ref{goal:alignment}, we co-design a unified whole-body controller used by both teleoperation and task-space target tracking, custom teleoperation hardware and interfaces, and a sensorized human data collection system. These components enable policies to efficiently learn coordinated and expressive whole-body behaviors on \robotname{}.

\textbf{Whole-Body Controller.}
We use a shared whole-body controller to maintain consistent motion characteristics across demonstrations collected with different devices, thereby reducing the mismatch between data collection and learned policy execution.

We use geometric fabrics~\cite{xie2021geometric,vanwyk2022geometric} to generate coordinated whole-body motion from task-space targets. Fabrics pair each local acceleration policy with a \emph{metric} that determines its relative influence and preferred directions of motion. A key benefit is their \emph{state-dependent metrics}, which allow motion priorities to adapt to the robot configuration, velocity, and targets as reachability or proximity to joint limits changes. The robot's kinematics map the local policies into a common configuration space, where they combine into a single acceleration field. COWBOY~\cite{yang2026cowboy} implements fabrics-based whole-body control for a mobile robot with one manipulation arm and an actuated camera arm. We extend fabrics-based whole-body control to a wheeled humanoid with two manipulation arms, an articulated torso, a vertical lift, and an actuated head.

\emph{Hand and head pose tracking.} Pose attractors draw the two grippers and the head toward desired positions and orientations, with damping near each target. These tasks act through the robot's full-body kinematics, so tracking a hand or head pose can coordinate base motion, lift height, and torso posture. Head targets specify both viewpoint and viewing direction: VR teleoperation derives the orientation from a hand-based look-at point (Equation~\ref{eq:main_head_gaze}), while learned policies predict head poses directly. Gripper apertures are controlled separately.

\emph{Reach-dependent base coordination.} A base-translation term favors arm motion when the hand targets are nearby and allows greater base participation as they move beyond reach. We approximate each arm's workspace by a sphere centered at its shoulder. When all active hand targets lie within these estimated workspaces, a large metric makes base acceleration more costly, encouraging the arms to accommodate the motion. As either target moves beyond its workspace radius, the metric decreases smoothly, allowing the pose attractors to recruit more base translation. Damping moderates base motion throughout this transition. This term lets the controller move continuously between local manipulation and coordinated base--arm reaching without a separate base-position target.

\emph{Posture and motion allocation.} Posture attractors favor a nominal body configuration while the hand and head targets are tracked. Leader-arm teleoperation can supply arm and torso posture references to preserve the operator's intended configuration. A separate motion-allocation metric assigns greater cost to lift and torso acceleration, biasing motion toward the arms while retaining whole-body flexibility. For base rotation, a yaw attractor follows a heading derived from the hand targets and desired viewing direction; a heading-dependent metric reduces unnecessary rotation near this heading. Together, these terms express weighted preferences for how the robot uses its redundant degrees of freedom.

\emph{Avoidance and motion generation.} Joint-limit repulsion steers joints away from their boundaries, while repulsion between link-attached spheres discourages self-collision. These behaviors contribute soft motion preferences to the combined controller. The resulting acceleration field is damped, bounded, and integrated at $200\,\mathrm{Hz}$ into joint-position and base-velocity references tracked by the low-level controllers. Appendix~\ref{sec:appendix_wbc} presents the mathematical formulation and controller parameters.

\begin{figure*}[t]
  \centering
  \includegraphics[
    width=\textwidth,
    keepaspectratio
  ]{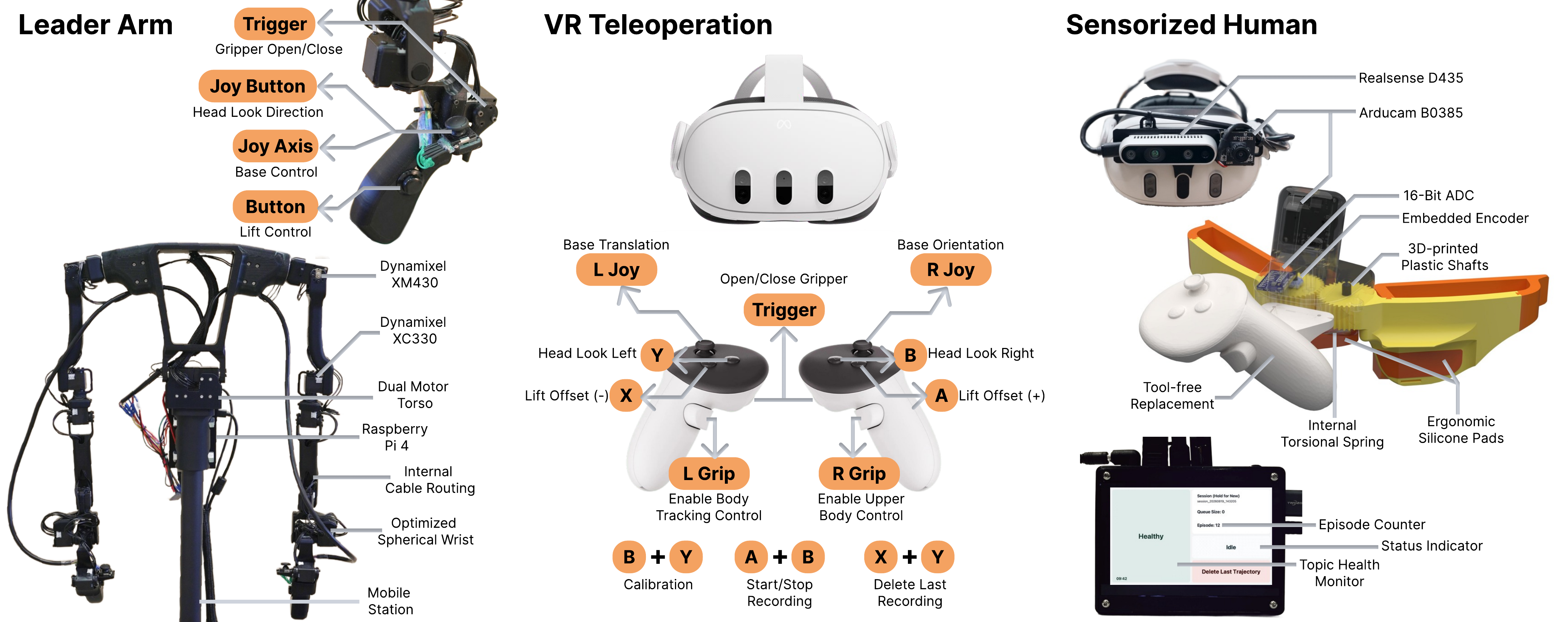}
  \caption{\textbf{Data collection interfaces.} \robotname{} supports three main forms of data collection. \textbf{Left:} A kinematically matched leader-arm station for precise bimanual teleoperation. \textbf{Center:} VR for untethered whole-body teleoperation. \textbf{Right:} Sensorized human data collection system for scaling robot-free demonstrations. All interfaces produce a common task-space representation used by the shared whole-body controller.}
  \label{fig:control_interfaces}
  \vspace{-4mm}
\end{figure*}

\textbf{Teleoperation Hardware and Interfaces}

To accommodate diverse users and tasks, we design two forms of teleoperation devices for the robot: a leader arm system for dexterous tabletop tasks and a VR teleoperation system for whole-body manipulation skills. Figure~\ref{fig:control_interfaces} presents our two teleoperation approaches.

\underline{\textit{Leader Arm System.}} For many fine-grained tasks, a leader arm system can be intuitive as it provides the teleoperator with a physically grounded representation of the robot \cite{wu2024gello}. We design a kinematically matched, 65\%-scale replica of the robot's torso and two 7-DoF arms, with integrated triggers for continuous gripper control. We omit the robot's actuated head from the replica and instead delegate gaze control to the whole-body controller, reducing the size and complexity of the interface without sacrificing gaze control.

The leader arm station is constructed primarily from 3D-printed parts and off-the-shelf Dynamixel servos, making the interface straightforward to reproduce and maintain in support of \ref{goal:hardware}. The joints are actuated and support features such as gravity and friction compensation, virtual joint limits, configurable null-space posture regulation, and bilateral force feedback~\cite{liu2025factr,oh2026factr2}. Gravity compensation reduces operator effort by counteracting the weight of the device, null-space regulation improves ergonomics by biasing each leader arm toward a comfortable configuration, and force feedback improves contact awareness by reflecting the robot's interaction forces back to the operator. Buttons integrated into the handles raise and lower the lift, while thumb joysticks command planar base translation and rotation.

A central design choice is that the measured leader-arm angles are not sent to the robot as joint-position targets. Instead, we use them in two complementary ways: we apply the leader configuration to the robot's kinematic model to compute gripper poses that serve as primary task-space attractors, and we provide the joint angles as a secondary nominal posture objective. The task-space attractors allow the whole-body controller to coordinate the arms with the torso, lift, and base while accounting for joint limits and self-collision avoidance. Because the gripper poses do not fully specify the configuration of the robot's redundant upper body, the nominal posture objective additionally biases the robot toward the operator's intended torso and elbow configuration. These objectives preserve both the task-level motion and the configuration-level intent of the operator to enable more expressive teleoperation for whole-body behaviors (\ref{goal:capability}). Hardware and controller details are provided in Appendix~\ref{sec:appendix_leader}.

\underline{\textit{VR Teleoperation System.}}  An untethered interface that allows the operator to move freely can make many whole-body tasks easier and more intuitive to teleoperate than with leader-arm systems. For example, the operator may need to reposition themselves for a better view of the scene or walk alongside the robot as it navigates through the environment. A VR-based interface supports this mobility while allowing the operator to continuously control the robot.

We use the Meta Quest 3/3S for its relatively low cost and self-contained 6-DoF tracking, making the system easier to reproduce and supporting \ref{goal:hardware}. The headset is worn around the operator's neck to provide body reference tracking, while the handheld controllers provide hand poses, continuous gripper commands, and auxiliary controls. Because the system is intended for local rather than remote teleoperation, the operator observes the robot and workspace directly instead of receiving streamed robot-camera images through the headset. This preserves the wide field of vision of human eyes and avoids latency-sensitive visual feedback, which can cause dizziness and disorientation during immersive teleoperation~\cite{xiong2025vision, chattha2020motion}. A custom OpenXR application based on \cite{wildersmith2024rfteleoperation} built in Unity transmits the headset and controller poses and inputs over the robot's private Wi-Fi network through a ROS-TCP bridge, and the robot-side interface updates the whole-body targets at $60\,\mathrm{Hz}$. Controller haptics confirm calibration and mode changes without requiring the operator to consult a display, allowing the user to focus on the task at hand.

The VR system provides headset and controller poses in the VR world frame, which must be mapped to the task-space attractors used by the robot's whole-body controller. We write $T_X^F$ for the pose of frame $X$ expressed in frame $F$. To reduce sensitivity to differences in operator anatomy and starting posture, teleoperation begins with a calibration in which the operator stands upright. The system records the headset pose, infers a human hip frame using a headset-to-hip offset, and records the robot's current head height.

During teleoperation, each controller is first expressed relative to the human hip and then transferred to the robot hip. We use the hip as the reference frame because of the strong correspondence between the human and robot upper bodies, providing a natural mapping of human upper-body motion to the robot. For the left and right controllers, $i\in\{L,R\}$, the corresponding gripper attractor is
\[
  \underbrace{T_{G_i}^{W}}_{\text{gripper attractor}}
  =
  \operatorname{Trans}_{W}(\delta_z\mathbf e_z)\,
  \underbrace{T_{R}^{W}}_{\text{robot hip}}
  \underbrace{\textcolor{black}{\left(T_{\bar U}^{V}\right)^{-1}}T_{C_i}^{V}}_{\text{controller relative to human hip}},
\]
where $V$ and $W$ are the VR and robot world frames, $U$ and $R$ are the human and robot hip frames, $C_i$ is a controller, and $G_i$ is its corresponding gripper attractor. {\color{black}For hand retargeting, $\bar U$ follows the estimated hip horizontally and in heading, while remaining level and at its calibrated height; the live height of $U$ is retained for head retargeting. Appendix~\ref{sec:appendix_quest} details these frames and their construction.} Here, $\mathbf e_z=[0,0,1]^\top$, and $\operatorname{Trans}_{W}(\delta_z\mathbf e_z)$ applies an operator-adjustable vertical-workspace offset along the robot-world vertical axis. Because the human and robot have different vertical workspaces, this offset lets the operator shift the gripper and head attractors together through the robot's lift range without maintaining a crouched or uncomfortable posture, supporting more efficient data collection \ref{goal:alignment}.

Since the headset is worn around the operator's neck, its pose cannot be mapped to the robot in the same way as the controller poses. Instead, we construct the head attractor poses in a decoupled fashion that preserves desired motions without directly using the headset pose. The horizontal position is predicted from the robot's kinematics using a candidate torso angle $\theta_{\mathrm{lean}}$ derived from the operator's forward lean:
\[
  \begin{bmatrix}
    x_{\mathrm{head}}^{*}\\
    y_{\mathrm{head}}^{*}
  \end{bmatrix}
  =
  \Pi_{xy}\!\left[
    \operatorname{FK}_{\mathrm{head}}^{W}
    \left(
      \mathbf q_t
      \big|_{q_{\mathrm{torso}}=\theta_{\mathrm{lean}}}
    \right)
  \right],
\]
where $\Pi_{xy}$ extracts the horizontal position from the robot head pose returned by forward kinematics. This maps the operator's lean angle through the robot's own kinematics rather than copying the operator's absolute head displacement, making the resulting head position consistent with the robot's geometry. Next, the vertical position follows the operator's change in hip height:
\[
  z_{\mathrm{head}}^{*}
  =
  z_{\mathrm{head},0}^{W}
  +
  \left(z_U^{V}-z_{U,0}^{V}\right)
  +
  \delta_z,
\]
where $z_{\mathrm{head},0}^{W}$ is the robot head height at calibration. To keep the manipulation region in view, we choose a look-at point above the midpoint of the two gripper targets and align the head's forward axis with it:
\begin{equation}
  \begin{aligned}
    \mathbf p_{\mathrm{look}}^{W}
    &=\frac{\mathbf p_{G_L}^{W}+\mathbf p_{G_R}^{W}}{2}
    +h_{\mathrm{look}}\mathbf e_z,\\
    \mathbf x_{\mathrm{head}}^*
    &=\frac{\mathbf p_{\mathrm{look}}^{W}-\widehat{\mathbf p}_{\mathrm{head}}^{W}}
    {\|\mathbf p_{\mathrm{look}}^{W}-\widehat{\mathbf p}_{\mathrm{head}}^{W}\|_2}.
  \end{aligned}
  \label{eq:main_head_gaze}
\end{equation}
Here, $\mathbf p_{G_i}^{W}$ is the position of gripper attractor $G_i$, $h_{\mathrm{look}}=0.20\,\mathrm m$ is the configured vertical look-at offset, and $\widehat{\mathbf p}_{\mathrm{head}}^{W}$ is the head position returned by the torso-adjusted forward kinematics above, before replacing its height with $z_{\mathrm{head}}^*$. The operator can instead select either hand, replacing the midpoint with that hand's target position. Projecting world up onto the plane perpendicular to $\mathbf x_{\mathrm{head}}^*$ determines the remaining head orientation; Appendix~\ref{sec:appendix_quest} gives the full rotation construction and its degenerate cases. The resulting gripper and head attractors are passed to the shared whole-body controller, which coordinates the motion of the robot's joints.

We provide two complementary base-control modes for different scenarios and compare them in Section~\ref{sec:control_interfaces_experiments}. \textit{VR Joystick} maps the controller thumbsticks to base-frame planar velocities, giving the operator explicit control over base motion. \textit{VR Body Tracking}, when engaged through a controller clutch, maps the translation and yaw of the human teleoperator onto the robot base. The robot can therefore follow the operator's body motion while remaining coordinated with the hand and head targets, which can be useful for whole-body tasks such as sweeping and opening doors. Both modes use the same whole-body controller and task-space target representation as the leader-arm interface and autonomous policy deployment.

\textbf{Sensorized Human Data Collection}

To enable more scalable data collection than teleoperation, we co-design a wearable human data collection system that enables \robotname{} to learn whole-body skills directly from humans performing tasks, without the need for a robot. The complete system uses off-the-shelf sensors and primarily 3D-printed parts, costs under \$1,000, and can operate untethered for approximately $2.5$--$3$ hours, consistent with the cost and reproducibility objectives of \ref{goal:hardware}. The complete system is shown in Figure~\ref{fig:control_interfaces}.

\begin{figure*}[t]
  \centering
  \includegraphics[
    width=\textwidth,
    keepaspectratio
  ]{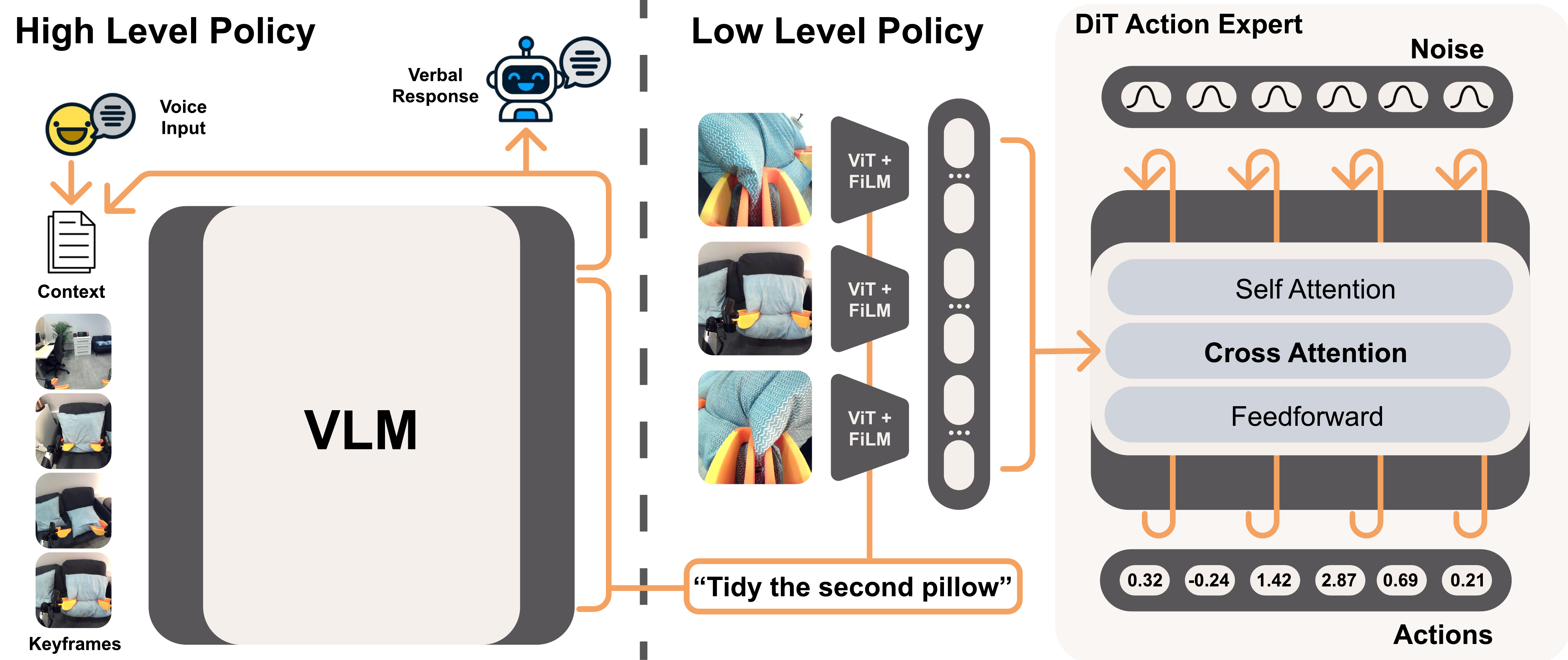}
  \caption{\textbf{Policy overview.} \textbf{Left:} The high-level policy processes voice commands, prior interaction context, and selected keyframes to determine the next subtask and generate a verbal response to the user. \textbf{Right:} The low-level policy conditions on the subtask description and visual observations to generate robot actions using a DiT-based action expert.}
  \label{fig:policy_architecture}
  \vspace{-4mm}
\end{figure*}

\underline{\textit{Hardware Co-Design.}} Central to our system is a co-designed gripper and VR headset that maintains strong alignment between the robot and the human (\ref{goal:alignment}). Each wearable gripper preserves the task-relevant finger geometry and wrist-camera
placement of the robot gripper, while minimally modifying the hand-facing structure for
better human ergonomics. Users place their thumb opposite the index and middle
fingers, forming a natural C-shaped pinch whose aperture directly actuates the
gripper. Silicone pads, elastic straps, and a support ridge distribute loads across the hand, enabling the manipulation of heavier everyday objects. A weak torsional spring assists reopening when the user relaxes their grasp, reducing the finger-extension effort required between grasps.

Because the wearable interface is human-powered, it does not require the same metal
transmission used on the robot. We instead integrate each gripper finger and its gear
into a single 3D-printed PETG part and use press-fit 3D-printed shafts, reducing part
count, cost, and mass and simplifying fabrication and replacement in support of
\ref{goal:hardware}. We use VR controllers for wrist tracking and attach them through their existing
battery-cover interface, allowing installation and
removal without tools. The resulting wearable gripper weighs only $386\,\mathrm{g}$, substantially less than comparable devices like the original UMI device at $780\,\mathrm{g}$~\cite{chi2024universal}, the AgileX PIKA Sense at $550\,\mathrm{g}$~\cite{agilex_pika_sense}, and the GenRobot DAS Gripper at $520\,\mathrm{g}$~\cite{genrobot_das_gripper}. Our lightweight and ergonomic interface reduces operator fatigue during repeated demonstrations, supporting sustained data collection in \ref{goal:alignment}.

We measure continuous gripper aperture using a resistive rotary encoder and a
16-bit ADC. Unlike vision-based aperture estimation~\cite{chi2024universal},
direct sensing remains available under occlusion and varying illumination and
provides low-latency measurements throughout fast motions, improving the fidelity of the recorded action signal in support of \ref{goal:alignment}. The sensor measures
aperture state rather than the operator's applied grasp action, so we also use the natural
mechanical compliance of the gripper to encode contact-dependent grasp intent. Specifically,
the wearable gripper uses a slightly softer membrane and contact skin than the
robot gripper. The two interfaces have approximately matching apertures in free
space, but contact causes greater deformation on the wearable interface and
therefore produces a smaller recorded aperture as the operator squeezes an
object. When transferred to the robot, this acts as a contact-dependent bias
toward further closure and a more secure grasp. Unlike a constant aperture
offset, it leaves free-space and fully open motion approximately unchanged.

We use a Meta Quest 3/3S to track the pose of the operator's head and both grippers.
Unlike VR teleoperation, the headset is worn normally during human data collection, enabling head motion to serve as
the robot's gaze target while providing a high-quality passthrough view of the
workspace. We reproduce the robot's head- and wrist-camera viewpoints on the
wearable system and calibrate the fixed transforms between the camera, the Quest controller and headset, and the attractor frames in the robot URDF used for whole-body control. Appendix~\ref{sec:appendix_human_calibration} describes these calibrations and the human--robot frame alignment. The Quest and all sensors connect to a Jetson Orin Nano carried in a custom enclosure
with a touchscreen. A portable $20{,}000\,\mathrm{mAh}$ battery powers the
compute, headset, cameras, and USB hub, so the complete system is self-contained and can move with the
operator without an external power or tracking setup.

\underline{\textit{Software \& Data.}} The software stack for human data collection is derived from the VR teleoperation stack, with major components such as the OpenXR Unity application, ROS~2 publishers and subscribers, and data collection code almost identical. During collection, users see a high-quality passthrough view while all menus remain hidden unless requested to reduce distractions. Distinct controller-haptic patterns
communicate calibration, recording start and stop, ongoing recording, and recording failures. In parallel, the touchscreen user interface reports the health and update rate of every recorded stream, allowing sensor dropouts to be
identified.

For each episode, we record time-stamped head and bimanual gripper poses,
continuous gripper apertures, the three RGB camera streams, and the head-mounted
RGB-D stream. Recording can be started and stopped directly from the controller,
so an operator does not need to interact with the touchscreen. Data storage, temporal synchronization, and training sample construction are detailed in Appendix~\ref{sec:appendix_data_pipeline}.

During lightweight post-processing, we transform the tracked head and gripper poses from the Quest convention into the task-space convention used by the robot, preserving a common action representation across human and robot demonstrations (\ref{goal:alignment}). We correct transient artifacts in the pose trajectories using heuristic jump detection and interpolation. Separately, we process the gripper-state signals using a Hampel filter~\cite{pearson2016generalized} for outlier rejection followed by Savitzky--Golay smoothing~\cite{savitzky1964smoothing}. Finally, we temporally rescale the trajectories to $0.5\times$ the demonstrated speed, which we found to improve tracking reliability.

For extensions of this sensorized human data system to more complex, dexterous tasks, we refer readers to our companion paper, Pixel-UMI~\cite{liu2026pixelumi}.

\subsection{Policy Learning}
\label{sec:policy}

Mobile manipulation presents a particularly challenging policy-learning
problem because a single task may combine coarse whole-body motion, such as
navigation, with precise manipulation,
such as grasping, insertion, or alignment. These behaviors unfold over horizons
substantially longer than those of typical tabletop tasks, increasing the effect
of control error accumulation and making policies more likely to encounter observations that
are poorly represented in the demonstration data. Moreover, open-ended tasks
may require several distinct skills whose order depends on the user's request
and the current state of the environment.

To realize broad everyday mobile-manipulation capability~\ref{goal:capability}, we design a
hierarchical system that separates task-level reasoning from continuous
control. A language-conditioned low-level policy executes reusable
mobile-manipulation skills, while a vision-language model (VLM) decomposes
open-ended user requests into supported skills and monitors their execution.
This hierarchy allows the low-level policy to specialize in closed-loop
visuomotor control while the VLM handles skill composition, progress tracking,
and interaction with the user, resulting in a powerful and general policy learning pipeline. Our policy architecture is shown in Figure~\ref{fig:policy_architecture}.

\textbf{Low-Level Multi-Task Policy.} We formulate low-level control as language-conditioned behavior cloning with action chunking~\cite{zhao2023learning}. At each timestep $t$, the policy receives an observation $\mathbf{o}_t$ and a short language instruction $\ell_t$, and predicts a horizon-$H$ action chunk

\[
  \mathbf{A}_t
  = \bigl[\mathbf{a}_{t+1},\ldots,\mathbf{a}_{t+H}\bigr]
  \in \mathbb{R}^{H\times d_a}.
\]

The policy therefore models the conditional distribution

$$\pi_\theta\left(\mathbf{A}_t \mid \mathbf{o}_t,\ell_t\right).$$

The observation $\mathbf{o}_t$ consists of synchronized RGB images from the head and two wrist cameras. The co-designed sensing and control interfaces described in Sections~\ref{sec:hardware} and \ref{sec:control_interfaces} align these observations and actions across robot teleoperation and sensorized human demonstrations, allowing both data sources to be used jointly for training. Because robot joint states have no direct counterpart in the human demonstrations, we omit robot-specific proprioception. Each demonstration trajectory is paired during data collection with a concise subtask instruction, such as ``put the vacuum back.''

We encode each camera view using a pretrained DINOv3 ViT-B backbone~\cite{simeoni2025dinov3}. For the head view, we retain the full token sequence to preserve spatial information about the broader scene~\cite{zhou2026patch}. For each wrist view, we retain only the class and register tokens, which provide compact representations of the local interaction region. We encode the language instruction with DistilBERT and mean-pool its output into a single embedding~\cite{sanh2020distilbertdistilledversionbert}. The visual tokens form the policy's context sequence, while the language embedding conditions this sequence through feature-wise linear modulation (FiLM)~\cite{perez2017filmvisualreasoninggeneral}. The visual backbones are initialized from pretraining and fine-tuned jointly with the policy, whereas the action generator is trained from scratch. Architecture and training details are provided in Appendix~\ref{sec:appendix_low_level_policy}.

The policy predicts relative pose trajectories \cite{chi2024universal} of the head and bimanual
end-effector attractors used by the whole-body controller, together with
continuous gripper apertures. To train our policies, we define an action space called \textit{Relative-base}. Instead of defining the reference frame as the pose of each attractor at the beginning of the chunk, we define our action space with respect to a gravity-aligned pseudo-base frame $B_t$ constructed from the head attractor at
the beginning of each chunk. The frame is placed on the ground directly below
the head and oriented using the head's horizontal facing direction. Head height,
roll, and pitch do not affect the frame. We express every future head and
end-effector target relative to this fixed chunk-start frame:
\[
  T_{j,t+k}^{B_t}
  =\left(T_{B_t}^{W}\right)^{-1}T_{j,t+k}^{W}.
\]
Here, $T_{B_t}^{W}$ is the chunk-start pseudo-base pose in world frame $W$, and
$j\in\{\mathrm{head},L,R\}$ identifies the head, left end effector, or right
end effector. Figure~\ref{fig:actionspace} gives a visual representation of this action space. Each pose is represented by its 3D position and a continuous 6D
rotation representation formed from the first two columns of its rotation
matrix. Together with the two continuous gripper commands, this gives a
29-dimensional action at each prediction step. The relative-base action space
avoids dependence on robot-specific joint kinematics, enabling sensorized human and robot demonstrations to share a common action representation and eliminating the action-space mismatch targeted by \ref{goal:alignment}. At deployment, the shared whole-body controller maps the predicted attractors to coordinated motions of robot joints.

\begin{figure*}[t]
  \centering
  \includegraphics[
    width=\textwidth,
    keepaspectratio
  ]{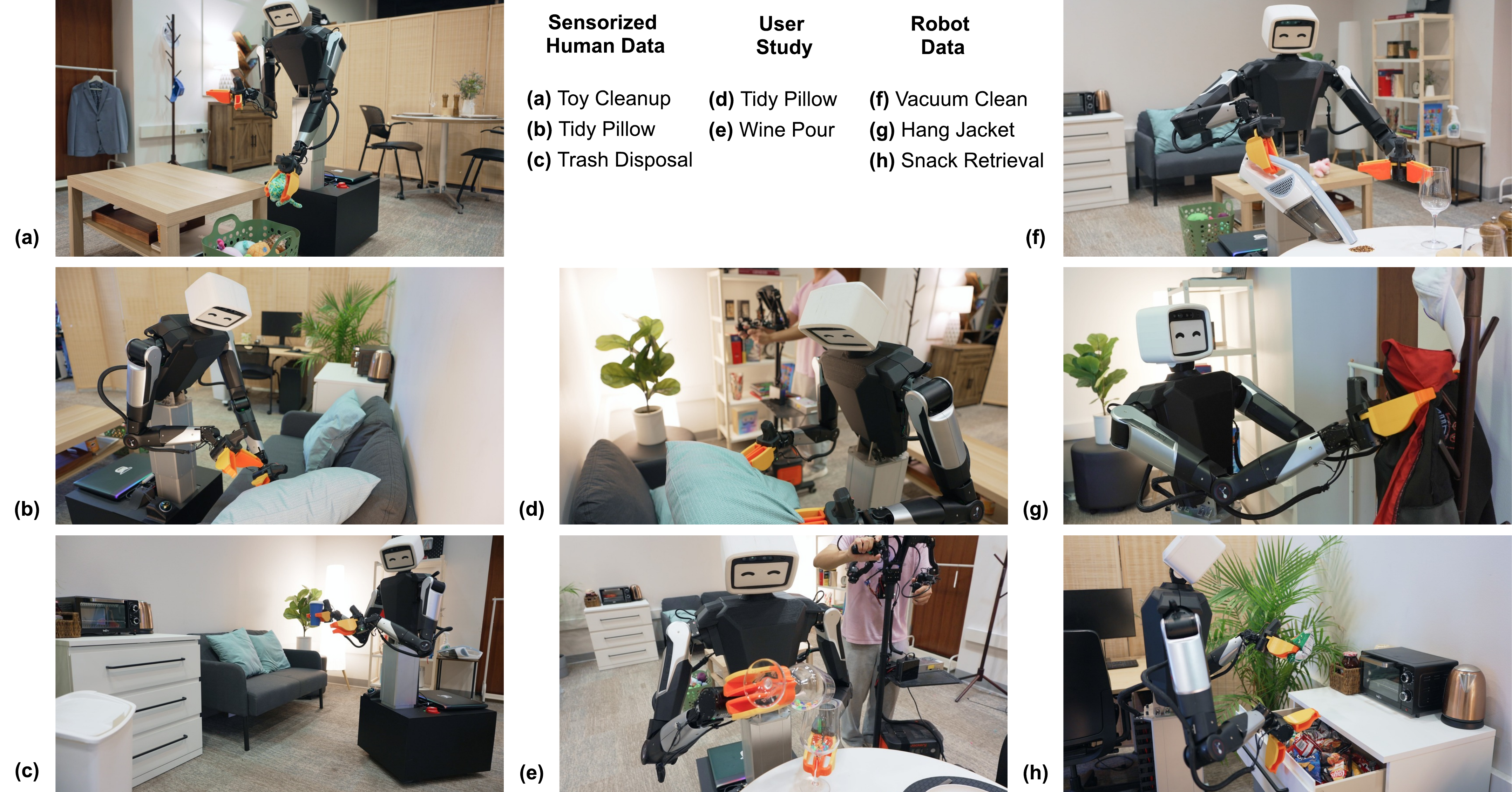}
  \caption{\textbf{Evaluation tasks.} We evaluate our robot hardware, data-collection interfaces, and policy-design choices across everyday mobile-manipulation tasks requiring precision, long-horizon execution, and whole-body coordination.}
  \label{fig:tasks}
  \vspace{-4mm}
\end{figure*}

We generate action chunks with an architecture we call \emph{FlowDiTX}, a DiT-style conditional
flow-matching transformer~\cite{peebles2023scalablediffusionmodelstransformers,lipman2023flowmatchinggenerativemodeling}. FlowDiTX
represents the noisy action chunk as $H$ action tokens. Each transformer block
first applies self-attention across the action tokens to model temporal
dependencies within the chunk, then applies cross-attention from the action
tokens to the language-conditioned visual context. To train the conditional
vector field, we sample a demonstration chunk $\mathbf{A}_t$, Gaussian noise
$\boldsymbol{\epsilon}\sim\mathcal{N}(\mathbf{0},\mathbf{I})$, and a flow time
$s\in[0,1]$. Let $\mathbf{c}_t$ denote the combined visual and language
context. We linearly interpolate between noise and data,
\[
  \mathbf{X}_s=(1-s)\boldsymbol{\epsilon}+s\mathbf{A}_t,
  \qquad
  \mathbf{u}_s=\mathbf{A}_t-\boldsymbol{\epsilon},
\]
and train the predicted velocity field $v_\theta$ using
\[
  \mathcal{L}_{\mathrm{FM}}
  =\mathbb{E}\!\left[
    \left\|v_\theta(\mathbf{X}_s,s,\mathbf{c}_t)-\mathbf{u}_s\right\|_F^2
  \right].
\]

At inference, we initialize $\mathbf{X}_0$ with Gaussian noise and numerically integrate from $s=0$ to $s=1$ using Euler's method to generate the predicted action chunk. The policy is executed in an asynchronous receding-horizon loop, where each new prediction is conditioned on the action prefix that will have been executed during the inference interval~\cite{black2025training}. This design reduces discontinuities between consecutive action chunks and compensates for inference latency, enabling smoother execution of the dexterous and whole-body behaviors targeted by \ref{goal:capability}. Appendix~\ref{sec:appendix_policy_deployment} describes the deployment pipeline and chunk alignment.

\textbf{High-Level Agent Harness.} The low-level policy provides closed-loop control for individual language-conditioned skills, but it is not designed to interpret open-ended requests, reason across multiple subtasks, or track progress over long horizons. We therefore place it within a VLM-based agent harness that separates task-level reasoning from physical execution. The harness operates in two phases: planning, which constructs a user-approved sequence of supported skills, and execution, which carries out this sequence while monitoring its progress. Separating these phases allows us to allocate more computation to planning and task-level reasoning before execution begins, enabling more reliable decomposition and management of long-horizon tasks, while favoring faster inference during execution monitoring in support of \ref{goal:capability}. We instantiate the high-level agent using Gemini 3.5 Flash.

During planning, the VLM receives the user request, the current head-camera observation, and, when replanning, the unfinished portion of the existing plan. It then constructs an ordered plan using a predefined skill library. Each skill is specified by a semantic description, a fixed instruction for the low-level policy, and visually verifiable success criteria. Restricting planning to this library grounds the VLM's decisions in the capabilities represented in the policy's training data and allows unsupported requests to be identified explicitly. The proposed plan is presented to the user for confirmation before execution begins.

After confirmation, the harness executes the plan one skill at a time. For each step, it maps the selected skill to its fixed low-level instruction, invokes the policy, and maintains the state of the overall plan. The VLM monitors execution using the skill's success criteria and a sparse visual history spanning the rollout. We construct this history from the observation at the start of the skill, a temporally intermediate keyframe selected to capture scene changes, a recent observation, and the latest camera image. This history serves as a compact visual memory, allowing the VLM to remember previous attempts, track meaningful progress, and determine whether the current skill has been completed while preserving fast inference. If progress repeatedly stalls or an unsafe condition is detected, the harness pauses execution and verbally requests assistance from the user. Users interact with the agent through speech and can confirm, pause, resume, stop, or revise the plan at any time. We use NVIDIA NeMo with the Parakeet Unified English 0.6B model for speech recognition and Piper for text-to-speech. Model configurations, prompts, and monitoring details are provided in Appendix~\ref{sec:appendix_high_level_policy}.

When paired, the high-level agent and low-level policy allow \robotname{} to compose reusable skills into longer-horizon behaviors while keeping physical execution grounded in the learned policy.

\section{EXPERIMENTS}

Having described the design of \robotname{}, we next evaluate how effectively the complete system realizes design goals~\ref{goal:hardware}--\ref{goal:capability}. Specifically, our experiments address the following questions:

\begin{enumerate}
  \item \textbf{Do the co-designed hardware components provide the mechanical capability required for everyday manipulation?} We evaluate how the robot's different whole-body degrees of freedom contribute to its task performance, conduct analyses on its workspace, and further measure the gripper characteristics.

  \item \textbf{Do the control interfaces enable intuitive and compatible collection of human and robot demonstrations?} We compare usability, collection efficiency, and ability to capture coordinated whole-body motion of different data collection methods.

  \item \textbf{Can we enable policies to learn reliable, long-horizon whole-body behaviors from demonstrations?} We explore different policy design choices including observation and action spaces to understand their effects on policy performance.

\end{enumerate}

These experiments evaluate the complete co-designed system, from its physical embodiment and data-collection interfaces to its policy-learning stack.

\subsection{Task Definitions}
\label{sec:tasks}

Several experiments use a common suite of tasks, summarized below by data-collection source and evaluation setting. Figure~\ref{fig:tasks}
illustrates these tasks. Additional task images and progress-based scoring rubrics are provided in Appendices~\ref{sec:appendix_task_details} and~\ref{sec:appendix_task_scoring}, respectively.

\textbf{Sensorized Human Data Tasks.}
We collect sensorized human demonstrations and train policies for the following tasks:

\begin{enumerate}
  \item \textbf{Toy Cleanup (103 demonstrations).}
    Starting from a randomized pose with the coffee table in view, the robot must approach the table while lowering its body, pick up a randomly placed toy, strafe toward a nearby basket, and deposit the toy inside. This task evaluates whole-body coordination and generalization across objects and locations.

  \item \textbf{Tidy Pillow (129 demonstrations).}
    Starting from a randomized location with the couch in view, the robot must navigate around obstacles such as a desk, coffee table, and toy basket before placing the couch pillows upright against the seat back. This task evaluates long-horizon navigation and bimanual manipulation.

  \item \textbf{Trash Disposal (189 demonstrations).}
    The robot begins in front of a couch with a coffee cup placed on the seat. After picking up the cup, it must back away, actively scan the room to determine whether the trash can is to its left or right, navigate to it, and dispose of the cup. This task evaluates the ability to learn active vision from human demonstrations.
\end{enumerate}

For Toy Cleanup and Tidy Pillow, we additionally collect 288 and 141 robot-teleoperated demonstrations, respectively, to study joint training with human and robot data.

\textbf{Robot Teleoperation Data Tasks.}
We collect robot-teleoperated demonstrations for the following tasks:

\begin{enumerate}
  \item \textbf{Vacuum Cleaning (133 demonstrations).}
    Starting from a randomized location, the robot must retrieve a handheld vacuum, turn it on, navigate to a dining table, vacuum spilled pepper flakes, return the vacuum to its storage location, and turn it off. This task evaluates long-horizon mobile manipulation and interaction with household appliances.

  \item \textbf{Hang Jacket (89 demonstrations).}
    The robot must locate and approach a person, wait to receive their jacket, transport it to a clothes rack, and hang it. This task evaluates navigation, human--robot interaction, and manipulation of deformable objects with varying materials.

  \item \textbf{Snack Retrieval (108 demonstrations).}
    The robot must navigate to a drawer, open it, retrieve a snack, close the drawer, locate the requesting person, and hand over the snack. This task evaluates articulated-object manipulation, human interaction, and careful handling of fragile objects such as bags of chips.
\end{enumerate}

\textbf{User Study Tasks.}
We use two tasks in the data-collection interface study:

\begin{enumerate}
  \item \textbf{Tidy Pillow.}
    We use the same task definition given above to evaluate whole-body navigation and bimanual coordination.

  \item \textbf{Wine-Glass Pouring.}
    Starting in front of a dining table, the robot must grasp one wine glass with each gripper and pour beads from the right glass into the left without spilling. This task evaluates precise grasping and coordinated bimanual manipulation.
\end{enumerate}

\subsection{Hardware Evaluation}
\label{sec:hardware_experiments}

We evaluate whether the co-designed robot embodiment provides the mechanical capabilities required for complex mobile manipulation while remaining low cost and simple in construction. Our evaluation consists of three experiments. First, we ablate the lift, torso, and head degrees of freedom to assess their contributions to learned task performance and the retargeting of human demonstrations. Second, we evaluate the alignment between \robotname{} and the human body in terms of reachable workspace and compare its workspace with those of open-source and commercial mobile manipulators. Lastly, we compare our custom gripper with a purpose-built commercial robot gripper.

\textbf{Robot Degree-of-Freedom Ablation.} We evaluate the contributions of the lift, torso, and head on the Toy Cleanup, Tidy Pillow, and Trash Disposal tasks defined in Section~\ref{sec:tasks}. These tasks collectively require a broad range of mobile-manipulation capabilities.

For each task, we train a single-task policy using the low-level policy architecture described in Section~\ref{sec:policy}. We then disable each of the lift, torso, and head degrees of freedom individually through the whole-body controller and evaluate the resulting autonomous policy. The same end-effector-based policy is used for all experiments. We conduct five trials for each experimental condition and report the task-progress scores.

\begin{figure}[!t]
  \centering
  \includegraphics[
    width=\columnwidth,
    keepaspectratio
  ]{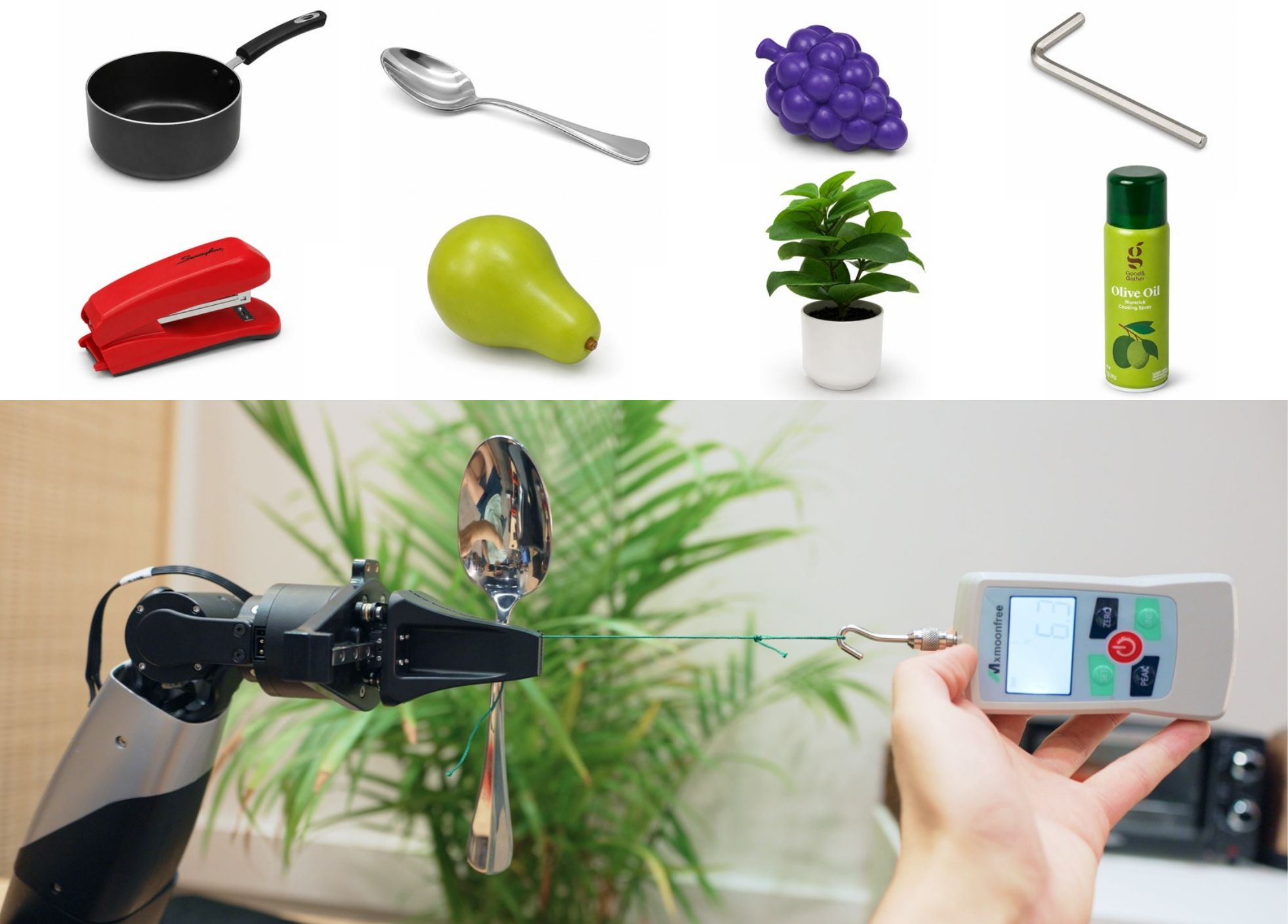}
  \caption{\textbf{Top:} Test objects for the gripper test. \textbf{Bottom:} Experimental setup showing a grasping test with the AgileX Piper gripper.}
  \label{fig:hardware_test}
  \vspace{-4mm}
\end{figure}

\begin{figure*}[t]
  \centering
  \includegraphics[
    width=\textwidth,
    keepaspectratio
  ]{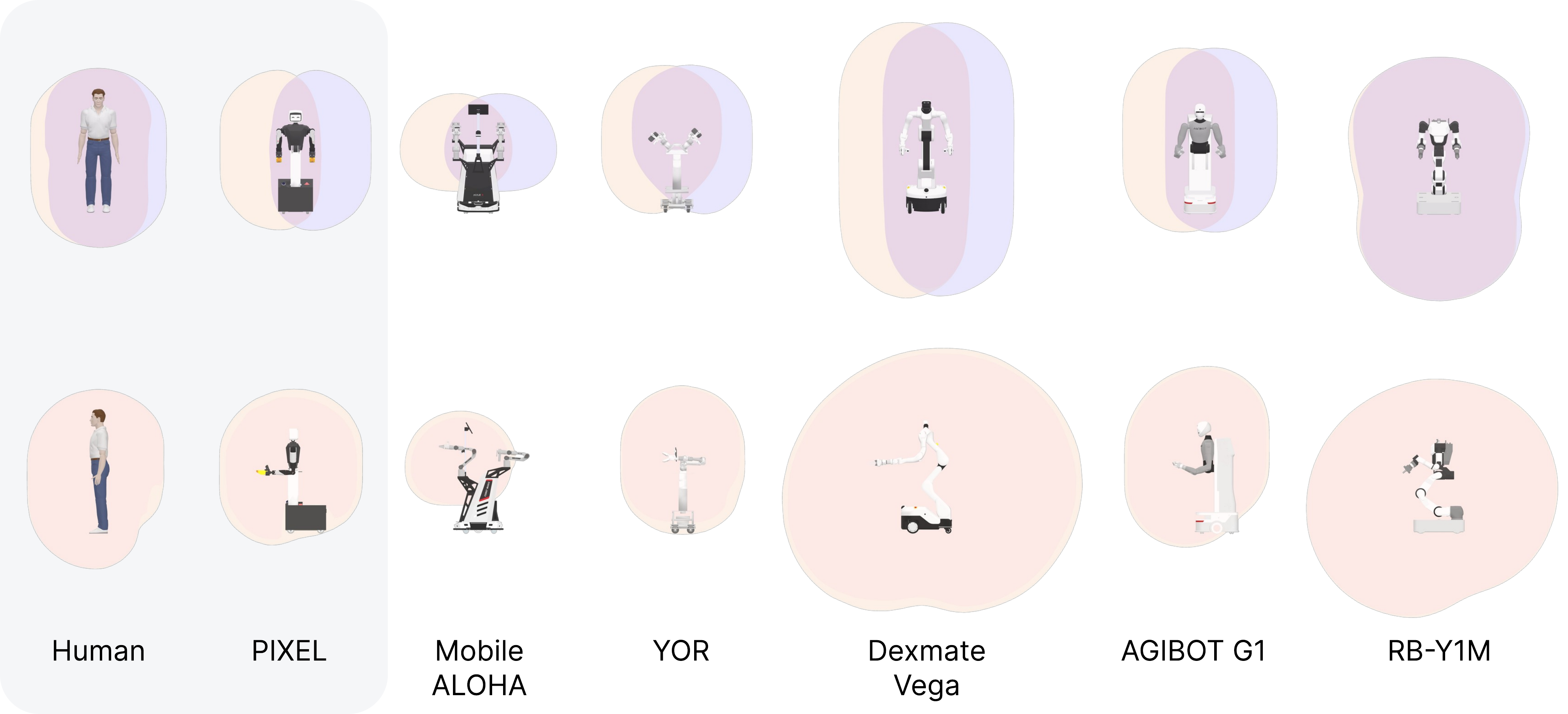}
  \caption{\textbf{Workspace comparison.} \robotname{} achieves the strongest human--robot workspace alignment among the evaluated platforms. Mobile ALOHA, YOR, and AGIBOT G1 cover a smaller fraction of the human workspace, while Dexmate Vega and RB-Y1M achieve broader reach through articulated lower bodies but substantially exceed human workspace volume.}
  \label{fig:workspace_compare}
  \vspace{-4mm}
\end{figure*}

\textbf{Workspace Analysis.} We evaluate the alignment between the workspace of \robotname{} and that of a canonical adult human model~\cite{human_description}. We additionally compare against YOR~\cite{anjaria2026yor}, AGIBOT-G1~\cite{agibot_g1}, Mobile ALOHA~\cite{fu2024mobile}, Rainbow Robotics RB-Y1M~\cite{rainbow_rby1}, and Dexmate Vega~\cite{dexmate_vega}. A visual comparison is shown in Figure~\ref{fig:workspace_compare}.

For each embodiment, we fix the base and sample configurations throughout its joint limits. Forward kinematics gives the Cartesian positions of the left and right end effectors, which we aggregate on a $25\,\mathrm{mm}$ voxel grid. For the human model, we sample the torso and arms across flat-footed postures from standing to squatting. All models share a ground plane and forward direction; \robotname{} is translated backward by $0.20\,\mathrm m$ for manual shoulder alignment. The comparison measures positional kinematic reachability subject to joint limits, without collision or stability filtering. Appendix~\ref{sec:appendix_workspace} details the sampling, discretization, and alignment procedure.

We report four complementary metrics to characterize workspace alignment. Let $H_L$ and $H_R$ denote the human left- and right-hand workspaces, and let $R_L$ and $R_R$ denote the corresponding robot workspaces. We define their combined workspaces as $H_U=H_L\cup H_R$ and $R_U=R_L\cup R_R$, where $|\cdot|$ denotes workspace volume.

\textit{Overall Reach Match} measures how closely the combined human and robot workspaces agree, regardless of which arm reaches a position. We count voxel centers within $\tau=50\,\mathrm{mm}$ of the other workspace as matches. Let $C_\tau^U$ be the fraction of human union voxels matched to the robot union and $P_\tau^U$ the reverse fraction. Their harmonic mean is
\begin{equation*}
  M_{\mathrm{overall}}
  =\frac{2C_\tau^U P_\tau^U}{C_\tau^U+P_\tau^U}.
\end{equation*}

\textit{Arm-Specific Reach Match} preserves arm identity, matching human left only to robot left and human right only to robot right. Pooling matched and total voxel counts across the two arms gives the directional fractions $C_\tau^A$ and $P_\tau^A$, with
\begin{equation*}
  M_{\mathrm{arm}}
  =\frac{2C_\tau^A P_\tau^A}{C_\tau^A+P_\tau^A}.
\end{equation*}
Both match scores use the same $50\,\mathrm{mm}$ tolerance; their full definitions are given in Appendix~\ref{sec:appendix_workspace}.

\textit{Human Reach Coverage} measures the fraction of the human workspace occupied by at least one robot arm, using exact voxel overlap without a tolerance:
\begin{equation*}
  C_{\mathrm{human}}
  =
  \frac{|H_U\cap R_U|}
  {|H_U|}.
\end{equation*}

\textit{Relative Workspace Volume} compares the total sizes of the robot and human workspaces:
\begin{equation*}
  \rho_{\mathrm{volume}}
  =
  \frac{|R_U|}
  {|H_U|}.
\end{equation*}
A value of $\rho_{\mathrm{volume}}=1$ indicates equal workspace volumes.

\textbf{Gripper Evaluation.} We compare our custom gripper with a commercial AgileX Piper gripper based on a linear-rail and linkage mechanism. The evaluation measures closing force and grasp retention across eight objects of varying shapes and sizes.

For each object, we conduct pull-out tests in both the vertical and horizontal directions. Fishing line is tied around the grasped object and connected to a force gauge, which applies an increasing load until the object is released or the applied force reaches $100\,\mathrm{N}$. We cap the measurement at $100\,\mathrm{N}$, beyond which further loading could damage the arm hardware, which has a rated load of approximately $30\,\mathrm{N}$. We perform three trials in each direction for every object, yielding 48 pull-out tests per gripper. An illustration of our experimental setup is in Figure~\ref{fig:hardware_test}.

\subsection{Data Collection Interface Evaluation}
\label{sec:control_interfaces_experiments}

We evaluate the four data-collection interfaces described in Section~\ref{sec:control_interfaces}: the Leader Arm system, VR Joystick, VR Body Tracking, and sensorized human data collection. Our evaluation examines the usability and collection efficiency of each interface.

\textbf{User Study.} We conduct a user study with 12 participants representing varying levels of prior teleoperation experience. Each participant performs the Tidy Pillow and Wine-Glass Pouring tasks defined in Section~\ref{sec:tasks} using all four interfaces.

Before beginning the study, participants receive instructions for each interface and 5 minutes to familiarize themselves with its controls. We randomize the order of the tasks and interfaces to reduce biases arising from learning and presentation order. Each participant performs two attempts per task with each interface, yielding 16 attempts per participant. We record task-completion time and administer a short survey to assess participants' preferences among the interfaces.

\textbf{Cross-Interface Data Compatibility.} We additionally evaluate whether sensorized human demonstrations and robot-teleoperated demonstrations can be used jointly for policy training. For the Toy Cleanup and Tidy Pillow tasks, we compare single-task policies trained using only sensorized human demonstrations with policies co-trained using both sensorized human and robot-teleoperated demonstrations. Data is uniformly sampled from the two datasets to ensure balanced training.

We evaluate each policy over 20 trials spanning four randomized task configurations and report the corresponding task-progress scores.

\begin{figure}[!t]
  \centering
  \includegraphics[
    width=\columnwidth,
    keepaspectratio
  ]{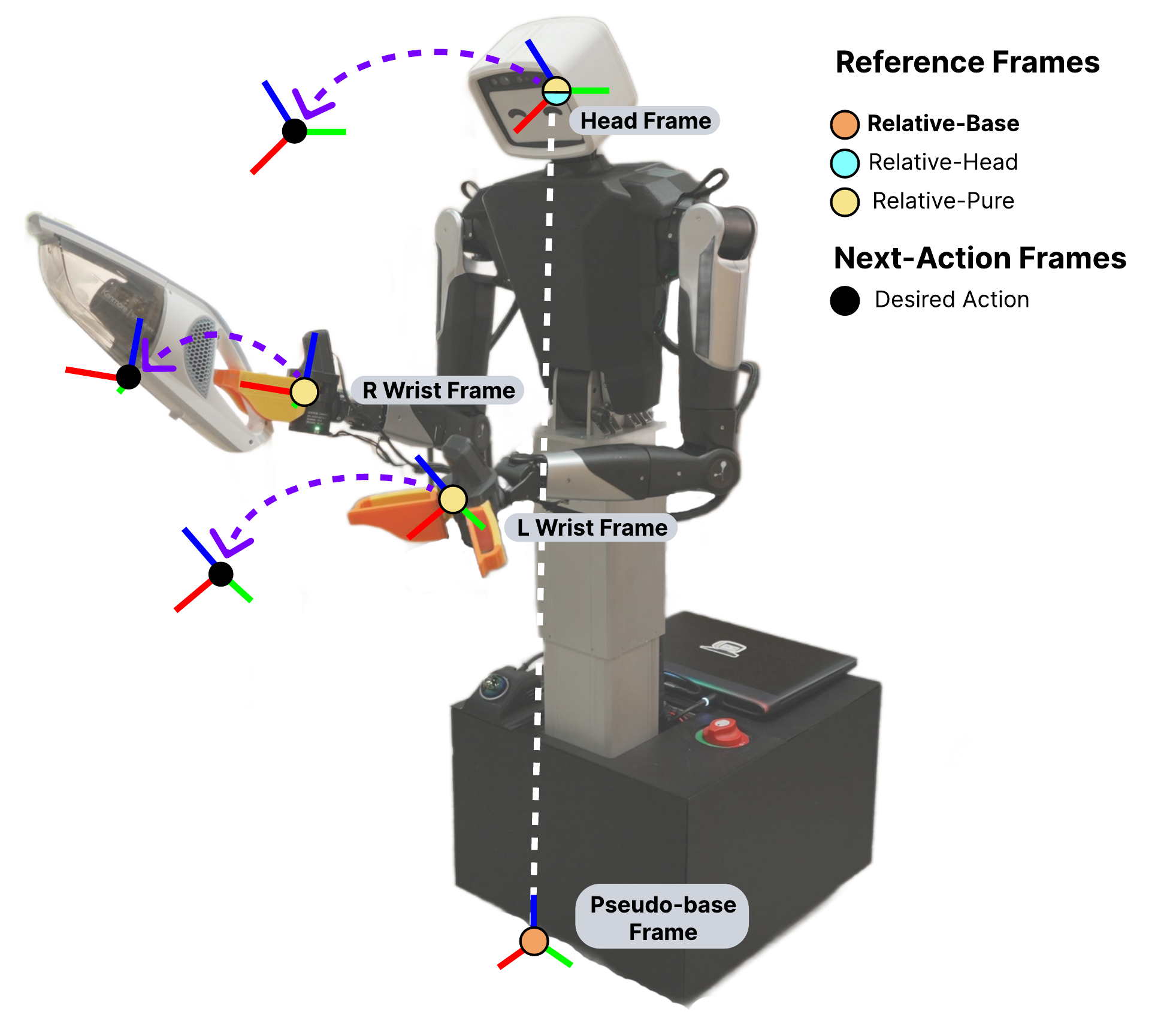}
  \caption{\textbf{Relative-base action space.} We express predicted head and bimanual end-effector trajectories in a shared pseudo-base frame constructed at the start of each action chunk. This representation preserves long-horizon precision and provides a common action space for sensorized human and robot demonstrations. \textit{Relative-pure} and \textit{Relative-head} reference frames are shown for comparison.}
  \label{fig:actionspace}
  \vspace{-4mm}
\end{figure}

\subsection{Policy Design Evaluation}
\label{sec:policy_design_experiments}
We evaluate how observation and action representations affect mobile-manipulation policy performance. Our final policy uses all three RGB cameras and the relative-base action space described in Section~\ref{sec:policy}. We assess the importance of these choices through three sets of ablations.

Unless otherwise specified, for each task and ablation condition, we train a separate single-task policy and evaluate it over 20 rollouts spanning four randomized task configurations. We report task-progress scores for all experiments.

\textbf{Observation-Space Ablations.} Our final policy receives images from the head camera and both wrist cameras. To determine whether all three views are necessary, we compare it with policies trained using only the head camera or only the wrist cameras. We conduct this experiment on both the sensorized human data tasks and the robot-teleoperation data tasks defined in Section~\ref{sec:tasks}.

\textbf{Action-Space Ablations.} For sensorized human data, we compare our relative-base action space with two alternative representations: \emph{relative-pure} and \emph{relative-head}. For robot teleoperation data, we compare our relative-base action space with joint-space actions. Figure~\ref{fig:actionspace} provides a visual comparison of all action spaces.

\underline{\textit{Sensorized human data:}} The relative-pure representation is equivalent to the relative action space introduced by UMI~\cite{chi2024universal}. Each predicted head and gripper trajectory is expressed relative to its own frame at the beginning of the action chunk. In contrast, relative-head expresses the head and both gripper trajectories in the head frame at the beginning of the chunk. This provides a shared reference frame for the gripper predictions but lacks the grounding of our pseudo-base frame. We evaluate the three action spaces on the sensorized human data tasks defined in Section~\ref{sec:tasks}.

\underline{\textit{Robot teleoperation data:}} We compare joint-space and end-effector-space policies on the robot-teleoperation data tasks defined in Section~\ref{sec:tasks}. The joint-space policies predict absolute targets for all robot joints together with a desired base velocity, producing a $23$-dimensional action space.

\section{RESULTS}

Numerical values underlying the result plots are provided in Appendix~\ref{sec:appendix_results}. For video results see: \href{https://pixel-robot.com/pixel/}{\texttt{https://pixel-robot.com/pixel}}.

\subsection{Hardware Experimental Results}

\textbf{\robotname{}'s set of degrees of freedom enables consistent performance across diverse mobile-manipulation tasks.}

\begin{figure*}[t]
  \centering
  \includegraphics[
    width=\textwidth,
    keepaspectratio
  ]{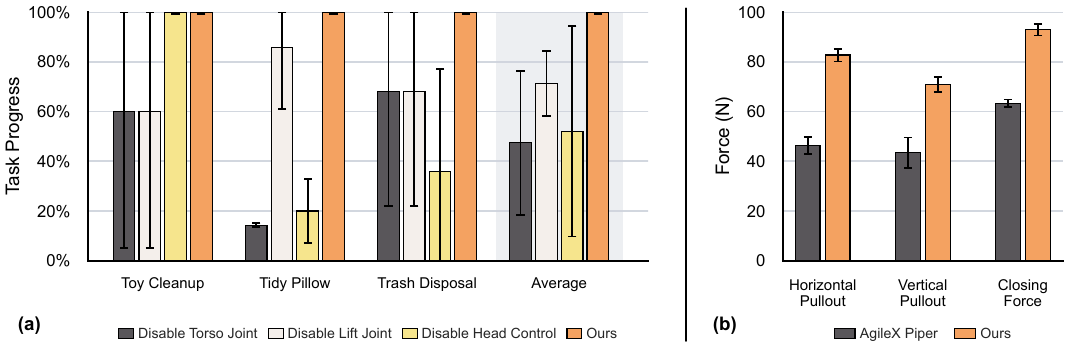}
  \caption{\textbf{(a)} We find that torso, lift, and head motion jointly support consistent task success across the evaluated tasks. \textbf{(b)} \robotname's gripper outperforms commercial alternatives such as the AgileX Piper gripper in grasping force and stability.}
  \label{fig:hardware_test_results}
\end{figure*}

As shown in Figure~\ref{fig:hardware_test_results}, the complete system achieved $100\%$ task progress on all three evaluated tasks. Averaged across tasks, progress decreased to $47.4\%$ when the torso was disabled, $71.2\%$ when the lift was disabled, and $52.0\%$ when head control was disabled.

The importance of each degree of freedom depended on the task. For example, Toy Cleanup retained $100\%$ progress without head control, suggesting that the visual encoder can tolerate the fixed camera viewpoint for some simple tasks. In contrast, Tidy Pillow progress decreased to $14.3\%$ without torso motion and $20.0\%$ without head control. The torso extends the robot's workspace when placing the pillow against the seat back, while active head control helps coordinate this whole-body motion. Trash Disposal progress similarly decreased to $36.0\%$ without head control. When collecting sensorized human demonstrations, humans naturally turn their heads to search for the trash can and keep it in view while approaching it. Disabling head control prevents the robot from realizing the head trajectories learned from this behavior, causing the camera viewpoint to diverge from the demonstrations and frequently preventing the robot from locating the trash can. Thus, although individual tasks may not require every degree of freedom, the complete embodiment enables consistent reproduction of the coordinated manipulation, navigation, and active-perception behaviors learned from sensorized human demonstrations.

\textbf{\robotname{} achieves the strongest alignment with the human workspace among the evaluated platforms.}
As shown in Table~\ref{tab:workspace_comparison}, \robotname{} achieves the highest Overall Reach Match at $90.99\%$, while covering $89.89\%$ of the human workspace with a Relative Workspace Volume of $1.11$.

Among platforms without articulated lower bodies, \robotname{} achieves the highest Overall Reach Match and Human Reach Coverage. YOR provides the closest comparison, achieving an Overall Reach Match of $89.80\%$ and the highest Arm-Specific Reach Match at $84.27\%$. However, YOR covers a smaller fraction of the human workspace ($79.65\%$), while AGIBOT-G1 and Mobile ALOHA achieve lower Overall Reach Matches of $75.22\%$ and $66.50\%$, respectively. These results suggest that \robotname{}'s combination of torso articulation and vertical lift provides broad human-aligned reach without requiring an articulated lower body.

Platforms with articulated lower bodies exhibit a different tradeoff. RB-Y1M and Dexmate both achieve $100\%$ Human Reach Coverage, but their workspaces are $3.02\times$ and $4.37\times$ the human workspace volume and produce substantially lower Overall Reach Matches of $54.65\%$ and $41.28\%$, respectively. Although their additional lower-body degrees of freedom expand fixed-base reach, much of the resulting workspace lies outside the human workspace. In contrast, \robotname{} provides a closer balance between coverage, workspace size, and spatial correspondence with human reach.

Because the base remains stationary in this analysis, these results isolate each platform's local manipulation workspace. During mobile manipulation, base repositioning can recover many otherwise unreachable positions. The coverage advantage provided by an articulated lower body may therefore be smaller when the full mobility of the platform is considered.

\begin{table}[t]
  \vspace*{8pt}
  \centering
  \caption{Human--robot workspace comparison. Reach-match scores use a $50\,\mathrm{mm}$ tolerance; coverage and relative volume use the original $25\,\mathrm{mm}$ voxel occupancy.}
  \label{tab:workspace_comparison}
  \footnotesize
  \setlength{\tabcolsep}{2.0pt}
  \renewcommand{\arraystretch}{1.10}

  \begin{tabular*}{\columnwidth}
    {@{\extracolsep{\fill}}lcccc@{}}
    \toprule
    \textbf{Robot}
    & \makecell{\textbf{Overall}\\\textbf{Match} $\uparrow$}
    & \makecell{\textbf{Arm-Spec.}\\\textbf{Match} $\uparrow$}
    & \makecell{\textbf{Human}\\\textbf{Coverage} $\uparrow$}
    & \makecell{\textbf{Relative}\\\textbf{Volume} $\rightarrow 1$} \\
    \midrule

    \textbf{\robotname{} (Ours)}
    & \cellcolor{metricorange!100}90.99\%
    & \cellcolor{metricorange!55}81.04\%
    & \cellcolor{metricorange!55}89.89\%
    & \cellcolor{metricorange!100}1.11 \\

    YOR
    & \cellcolor{metricorange!55}89.80\%
    & \cellcolor{metricorange!100}84.27\%
    & \cellcolor{metricorange!14}79.65\%
    & \cellcolor{metricorange!100}0.89 \\

    AGIBOT-G1
    & \cellcolor{metricorange!28}75.22\%
    & \cellcolor{metricorange!28}74.44\%
    & \cellcolor{metricorange!28}80.71\%
    & \cellcolor{metricorange!55}1.30 \\

    Mobile ALOHA
    & \cellcolor{metricorange!14}66.50\%
    & \cellcolor{metricorange!14}57.47\%
    & \cellcolor{metricorange!3}46.00\%
    & \cellcolor{metricorange!28}0.52 \\

    RB-Y1M
    & \cellcolor{metricorange!7}54.65\%
    & \cellcolor{metricorange!7}52.29\%
    & \cellcolor{metricorange!100}100.00\%
    & \cellcolor{metricorange!7}3.02 \\

    Dexmate
    & \cellcolor{metricorange!3}41.28\%
    & \cellcolor{metricorange!3}46.79\%
    & \cellcolor{metricorange!100}100.00\%
    & \cellcolor{metricorange!3}4.37 \\

    \bottomrule
  \end{tabular*}
  \par\vspace{-8pt}
\end{table}

\textbf{\robotname{}'s low-cost gripper outperforms the commercial AgileX Piper gripper in grasp retention and closing force.}
Our gripper resisted $82.7 \pm 2.6\,\mathrm{N}$ during horizontal pull-out tests, compared with $46.4 \pm 3.4\,\mathrm{N}$ for the AgileX Piper gripper, an improvement of $78\%$. During vertical pull-out tests, it resisted $71.0 \pm 3.0\,\mathrm{N}$, compared with $43.4 \pm 6.1\,\mathrm{N}$, an improvement of $63\%$. Its measured closing force was $92.9 \pm 2.4\,\mathrm{N}$, compared with $63.3 \pm 1.5\,\mathrm{N}$, an improvement of $47\%$. Thus, across the three tests, our gripper produced $1.47\times$--$1.78\times$ the force of the commercial comparison gripper despite using the same motor for actuation.

Moreover, our gripper reached the $100\,\mathrm{N}$ experimental safety limit in half of the pull-out trials, meaning that the object could not be removed before reaching the measurement cap. Because these measurements are capped, the reported pull-out forces conservatively underestimate the gripper's actual retention force in these trials. We attribute this performance to the combination of the compliant membrane, silicone contact surfaces, and rigid supporting structure, which allows the gripper to conform to objects while maintaining a strong grasp.

\begin{figure*}[t]
  \centering
  \includegraphics[
    width=\textwidth,
    keepaspectratio
  ]{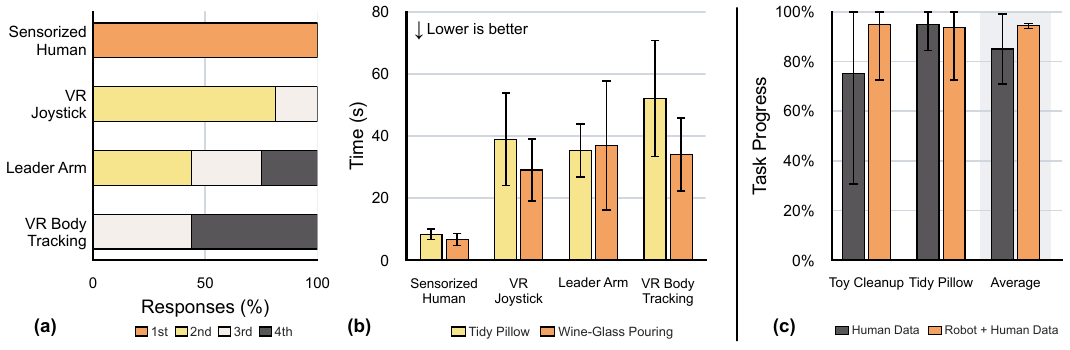}
  \caption{\textbf{(a)} User preferences obtained through a survey on different data collection methods.
    \textbf{(b)} Mean completion times for two user-study tasks using different data collection interfaces.
  \textbf{(c)} Our shared interface and controller design enables co-training to improve performance.}
  \label{fig:interface_design_results}
\end{figure*}

\subsection{Data Collection Interface Experimental Results}

\textbf{VR Joystick provides the strongest all-around teleoperation interface, while sensorized human data collection is unanimously preferred and substantially faster than teleoperation.}

\underline{\textit{Comparing teleoperation approaches.}} As shown in Figure~\ref{fig:interface_design_results}, participants preferred VR Joystick among the three teleoperation interfaces. It was most frequently ranked as the best teleoperation option, while Leader Arm occupied an intermediate position with a more balanced spread across rankings. VR Body Tracking ranked lowest, with most participants listing it as their least-preferred interface.

This preference ranking was partially reflected in mean task-completion time. Leader Arm was slightly faster than VR Joystick on Tidy Pillow ($35.3\,\mathrm{s}$ versus $39.0\,\mathrm{s}$), whereas VR Joystick was faster on Wine-Glass Pouring ($29.1\,\mathrm{s}$ versus $36.9\,\mathrm{s}$). VR Body Tracking had the longest mean completion time on Tidy Pillow, for which accurate base positioning was particularly important. Although Leader Arm did not provide a consistent completion-time advantage under the evaluated conditions, neither task was designed to isolate the benefits of physical grounding or force feedback. We hypothesize that Leader Arm may provide a larger advantage in contact-rich tasks requiring sustained contact and accurate force application, such as operating stiff handles, opening doors, or performing constrained insertion.

Participants' qualitative feedback provides a possible explanation for these results. Users preferred the explicit joystick-based control of the mobile base available in VR Joystick and Leader Arm over mapping their body motion to the robot base through the whole-body controller. They reported that the joystick enabled more precise base placement, which may explain why VR Body Tracking was both less preferred and slower on Tidy Pillow, a task requiring substantial base movement.

These results suggest that VR Joystick provides the strongest overall balance of user preference and completion time among the evaluated teleoperation interfaces, supporting its use as the default interface for \robotname{}.

\underline{\textit{Sensorized human data vs. teleoperation.}}
As shown in Figure~\ref{fig:interface_design_results}, every participant ranked Sensorized Human first for both tasks. This preference was reflected in task-completion time. Sensorized Human required only $8.4 \pm 1.7\,\mathrm{s}$ for Tidy Pillow and $6.7 \pm 1.9\,\mathrm{s}$ for Wine-Glass Pouring. By comparison, the three teleoperation interfaces required $35.3$--$52.1\,\mathrm{s}$ and $29.1$--$36.9\,\mathrm{s}$ for the two tasks, respectively. Sensorized Human therefore achieved a $4.2\times$--$6.2\times$ speedup over teleoperation across the interface--task comparisons. Completion times were also substantially less variable, with standard deviations of only $1.7$--$1.9\,\mathrm{s}$ compared with $8.5$--$20.8\,\mathrm{s}$ for teleoperation, indicating more consistent performance across participants with varying levels of prior experience.

Participants' qualitative feedback offered a possible explanation for these results. Sensorized Human required little acclimation because users could perform the tasks directly using their own whole-body motion rather than translating their intent through an intermediate robot-control interface. Its preference, speed, and consistency advantages held across both the whole-body Tidy Pillow task and the dexterous Wine-Glass Pouring task, supporting direct, embodiment-aligned human demonstrations as an efficient approach to scalable mobile-manipulation data collection.

\begin{figure*}[t]
  \centering
  \includegraphics[
    width=\textwidth,
    keepaspectratio
  ]{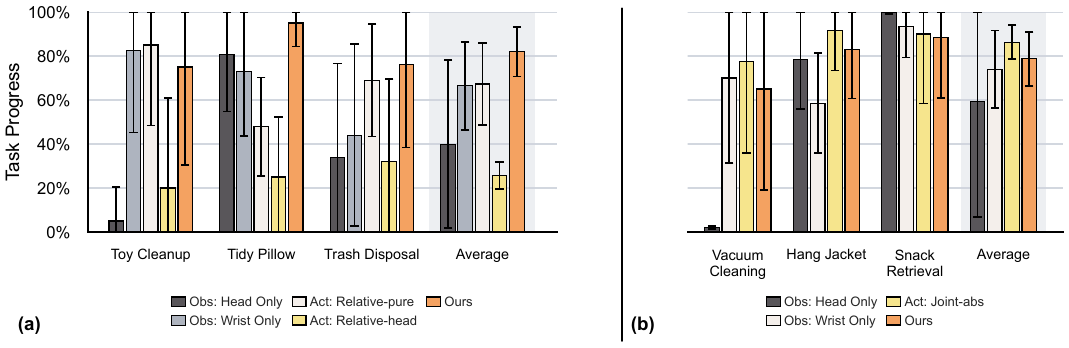}
  \caption{\textbf{(a)} Comparison of observation- and action-space design choices for policy training with \textbf{sensorized human data}.
  \textbf{(b)} Comparison of observation- and action-space design choices for policy training with \textbf{robot teleoperation data}.}
  \label{fig:policy_design_results}
\end{figure*}

\textbf{Human--robot alignment enables joint training on sensorized human and robot-teleoperated demonstrations.}
\mbox{Figure~\ref{fig:interface_design_results}} further shows that balanced co-training increased average task progress from $85.0\%$ using sensorized human data alone to $94.3\%$ across Toy Cleanup and Tidy Pillow. This improvement was driven by Toy Cleanup, where progress increased from $75.0\%$ to $95.0\%$. On Tidy Pillow, performance remained comparable, changing from $95.0\%$ to $93.6\%$. These results suggest that the shared observation and action spaces, together with the whole-body controller, place sensorized human and teleoperated demonstrations in a compatible representation space, allowing the system to combine efficiently collected human behavior with robot-specific data to improve policy performance.

\subsection{Policy Design Experimental Results}

\textbf{Complementary head and wrist views enable both fine-grained manipulation and active perception.}\
As shown in Figure~\ref{fig:policy_design_results}, using all three cameras produced the highest average task progress for both data sources. For policies trained with sensorized human demonstrations, the complete observation space achieved $82.0\%$ average progress, compared with $66.5\%$ using only the wrist cameras and $39.9\%$ using only the head camera. The same trend held for robot-teleoperated data, for which the complete observation space achieved $78.7\%$ average progress, compared with $74.0\%$ and $59.5\%$ for the wrist-only and head-only configurations, respectively.

The individual tasks illustrate why both perspectives are needed. Head-only policies achieved just $5.0\%$ progress on Toy Cleanup and $0\%$ on Vacuum Cleaning, consistent with the loss of close-range visual information needed to align grasps and interact with small controls such as the vacuum power button. Removing the head camera also prevents the policy from reliably reproducing active-perception behaviors present in the sensorized human demonstrations. During Trash Disposal, human demonstrators naturally turn their heads while backing away from the couch to search for and locate the trash can. Although the corresponding head motion remains part of the predicted action, a wrist-only policy does not observe the head-camera images that give this behavior its visual context. During Hang Jacket, the garment can cover both wrist cameras as it is transferred from the person to the robot. A wrist-only policy can therefore become completely visually occluded at a critical point in the task, whereas the head camera retains a view of the person, garment, and surrounding scene. Thus, the complete observation space is not necessary for every individual task, but it provides the most consistent performance across behaviors requiring close-range manipulation, active visual search, and robustness to camera occlusion.

\textbf{A shared, ground-anchored action frame improves long-horizon coordination while remaining compatible with both human and robot data.}\
On the sensorized-human tasks, our relative-base action space achieved $82.0\%$ average task progress, outperforming relative-pure at $67.3\%$ and relative-head at $25.7\%$. The difference was most pronounced on Tidy Pillow, where relative-base achieved $95.0\%$ progress, compared with $47.9\%$ for relative-pure and $25.0\%$ for relative-head. Relative-pure remained effective on shorter or less tightly coordinated tasks, reaching $85.0\%$ on Toy Cleanup and $69.0\%$ on Trash Disposal. However, because the head and two end effectors are independently referenced at the beginning of each action chunk, we observed drift and loss of bimanual coordination during longer-horizon behaviors. Relative-head provides a shared reference frame, but it couples the gripper trajectories to errors in the predicted head pose. We observed that vertical head-pose drift consequently propagated to both end effectors and frequently caused task failure. In contrast, relative-base expresses all three trajectories in a common chunk-start frame while preserving ground-anchored height and orientation, preventing head-prediction errors from redefining the gripper reference frame.

On the robot-teleoperation tasks, absolute joint-space actions achieved $86.3\%$ average progress, compared with $78.7\%$ for relative-base end-effector actions. We therefore do not conclude that end-effector actions are intrinsically superior to joint-space control. Rather, we choose relative-base as it retains strong robot performance while providing a representation that is directly available from both sensorized human and robot-teleoperated demonstrations.

\section{CONCLUSION \& LIMITATIONS}

We presented \robotname{}, an open-source, co-designed wheeled humanoid and integrated robot-learning stack spanning hardware, whole-body control, data collection, and policy learning. Rather than developing these layers in isolation, \robotname{} co-designs them around a central goal of enabling broad autonomous mobile manipulation. Our evaluations show that this full-stack approach supports dexterous, long-horizon mobile manipulation while keeping the system low-cost, reproducible, and safe to operate in human environments.

Several limitations point to important directions for future work. First, \robotname{}'s wheeled, low-cost embodiment trades some mobility and dexterity for simplicity and safety: it cannot traverse stairs or match the range of motion and manipulation capabilities of a full humanoid with dexterous hands. Future work should systematically characterize these tradeoffs and explore how improved affordable actuators and mechanisms can expand capability without compromising reproducibility or safety. Second, the sensorized human interface records head and gripper trajectories but does not capture uninstrumented body motions or contacts, such as bracing a door with an elbow, and lacks tactile sensing. Richer full-body tracking, action estimation, and multimodal sensing could preserve more of the expressivity of human demonstrations. Finally, the high-level agent's ability to compose learned skills has not been systematically evaluated. Because it operates over a predefined skill library, it cannot execute unsupported skills and offers limited compositional generalization. Expanding this repertoire, improving hierarchical policy architectures, and evaluating them on longer-horizon tasks remain important directions.

Ultimately, \robotname{} provides an open foundation for advancing hardware, data, and learning together.

\section*{ACKNOWLEDGMENT}

We would like to thank Philip Han and Ray Liu for their help with data collection. We would also like to thank Kenneth Shaw, Arthur Allshire, Ritvik Singh, Andrew Wang, Zheyuan Hu, Peiqi Liu, Sri Anumakonda, Steven Oh, Adam Kan, Haoran Huan, Jiahui Yang, Hengkai Pan, Lili Chen, Mihir Prabhudesai, Wenli Xiao, and Tairan He for their thoughtful discussions.  We would like to thank all our participants in the user study for their time and feedback. This work was supported by AFOSR FA9550-23-1-0747, ONR MURI N00014-22-1-2773 and ONR MURI N00014-24-1-2748.

\bibliographystyle{IEEEtran}
\bibliography{references}

\clearpage
\appendix
\label{sec:appendix}

\begingroup
\color{black}
\subsection{Base Control}
\label{sec:appendix_base}

\textbf{Swerve kinematics.}
The base controller maps a desired body-frame planar velocity $\boldsymbol{\nu}=[v_x,v_y,\omega]^\mathsf{T}$ into steering and drive commands for four swerve modules. For a module whose steering axis lies at $(x_i,y_i)$ relative to the base origin, the corresponding planar velocity is
\begin{equation}
  \mathbf{v}_i = \mathbf{J}_i\boldsymbol{\nu},
  \qquad
  \mathbf{J}_i =
  \begin{bmatrix}
    1 & 0 & -y_i \\
    0 & 1 & x_i
  \end{bmatrix}.
  \label{eq:appendix_base_jacobian}
\end{equation}
This combines chassis translation with the tangential velocity induced by rotation. With zero caster offset, the steering axis coincides with the wheel-center projection onto the ground. The desired steering angle and wheel speed are therefore
\begin{equation}
  \theta_i^\star=\operatorname{atan2}(v_{i,y},v_{i,x}),
  \qquad
  \dot\phi_i^\star=\frac{\|\mathbf{v}_i\|_2}{r},
  \label{eq:appendix_swerve_targets}
\end{equation}
where $r$ is the wheel radius, $\theta_i$ is the steering angle relative to the chassis, and $\phi_i$ is the wheel rotation angle.

\textbf{Relationship to offset-caster designs.}
To see the effect of caster offset, consider a wheel displaced from its steering axis by $(b_x,b_y)$ in coordinates aligned with its rolling direction. Under ideal rolling without lateral slip,
\begin{equation}
  \mathbf{J}_i\boldsymbol{\nu}
  = R(\theta_i)
  \begin{bmatrix}
    r & b_y \\
    0 & -b_x
  \end{bmatrix}
  \begin{bmatrix}
    \dot\phi_i \\
    \dot\theta_i+\omega
  \end{bmatrix},
  \label{eq:appendix_caster_offset}
\end{equation}
where $R(\theta_i)$ is a planar rotation and $\dot\theta_i+\omega$ is the absolute steering rate. The $2\times2$ velocity map has determinant $-rb_x$: a nonzero longitudinal offset makes it full rank, allowing steering to generate velocity perpendicular to the rolling direction. TidyBot++ introduces this offset to obtain holonomic motion without first aligning the wheels with the desired travel direction~\cite{wu2024tidybotopensourceholonomicmobile}.

For our zero-offset design, $b_x=b_y=0$, so the steering column vanishes and the module velocity map has rank one. Steering changes the rolling direction but contributes no direct translation in the ideal model; arbitrary travel directions therefore require steering alignment. We choose this tradeoff to decouple steering from offset-induced chassis motion and retain unmodified commercial modules, simplifying both the mechanical design and the direction-and-speed control law. The resulting base is omnidirectional, with steering transients handled as follows.

\textbf{Steering alignment and drive coordination.}
We reverse wheel rotation and shift the desired steering angle by $\pi$ whenever doing so reduces steering travel, keeping the angular error within $90^\circ$. A proportional steering controller tracks the selected angle through bounded steering-rate commands. During alignment, the signed wheel-speed target is scaled by $\max(0,\cos e_i)$, where $e_i$ is the remaining steering error. This projects the desired velocity onto the current rolling direction and reduces unintended motion while the wheel turns.

\textbf{Low-speed hysteresis.}
Near rest, small velocity fluctuations can produce large changes in the desired steering direction. We therefore enter a hold state when the commanded translational speed and absolute yaw rate fall below $0.01\,\mathrm{m/s}$ and $0.07\,\mathrm{rad/s}$, respectively. Hold is released when either reaches its larger threshold of $0.05\,\mathrm{m/s}$ or $0.08\,\mathrm{rad/s}$. During hold, steering-rate and wheel-speed commands are zero. These separate entry and release thresholds suppress repeated switching and steering adjustments near rest.

\textbf{Base-state estimation.}
Measured steering angles and wheel speeds provide module rolling velocities, from which we estimate chassis velocity by solving the stacked Jacobian equations in the least-squares sense. Integrating this velocity yields wheel odometry. When available, SLAM pose measurements are fused with wheel-derived velocity in a planar Kalman filter to estimate the base pose.
\par
\endgroup

\begingroup
\color{black}
\subsection{Lift Design and Control}
\label{sec:appendix_lift}

\textbf{Mechanism and sizing.}
A Kraken X60 motor drives the telescopic lead-screw column through a chain linking a 10-tooth motor sprocket and a 36-tooth screw sprocket ($G=3.6$). Let $L$ denote the effective screw lead, $\omega_m$ the motor angular velocity, and $F$ the axial lifting load. The lift speed and required steady-lifting torque are
\begin{equation}
  \dot z=\frac{L\omega_m}{2\pi G},
  \qquad
  \tau_m\approx\frac{FL}{2\pi\eta G},
  \label{eq:appendix_lift_sizing}
\end{equation}
where $\eta$ accounts for the overall transmission losses. The estimated supported mass is $26\,\mathrm{kg}$, including $6\,\mathrm{kg}$ of carried objects. A sizing factor of safety of approximately $1.54$ gives a $40\,\mathrm{kg}$ design load, corresponding to $F=40\times9.81=392.4\,\mathrm{N}$. With $L=3\,\mathrm{mm}$ and an assumed efficiency $\eta=0.30$, the required motor torque is approximately $0.17\,\mathrm{N\,m}$. The nominal lift speed is $75\,\mathrm{mm/s}$, corresponding to a motor speed of $5400\,\mathrm{rpm}$. At this design point, the manufacturer's FOC curve indicates approximately $0.65\,\mathrm{N\,m}$~\cite{wcp_kraken_performance}, about $3.7$ times the calculated lifting torque.

\textbf{Calibration and position control.}
The lift is controlled as a single prismatic joint. Desired positions from the laptop are converted onboard to motor targets using an encoder-to-height calibration referenced to both end switches. Commands remain within the calibrated travel, and either switch stops the motor. Calibration is saved between runs.

\textbf{Passive load holding.}
The column holds the upper body stationary when unpowered. Self-locking lead screws resist backdriving through thread friction, providing passive support during power loss.
\par
\endgroup

\begingroup
\color{black}
\begingroup
\color{black}
\subsection{Whole-Body Control with Geometric Fabrics}
\label{sec:appendix_wbc}

\subsubsection{Configuration and task representation}
We formulate whole-body control as the composition of local motion policies using geometric fabrics~\cite{xie2021geometric,vanwyk2022geometric}. Each policy specifies a desired acceleration and a metric that determines its influence on the combined motion. State-dependent metrics allow this influence to vary with reachability, joint configuration, and proximity to self-collision. Given desired poses $T_{G_L}^{W,\star},T_{G_R}^{W,\star},T_{\mathrm{head}}^{W,\star}\in SE(3)$ for the two grippers and head in the robot world frame $W$, together with a nominal posture, the controller generates coordinated base and upper-body motion. We write the robot configuration as
\begin{equation}
  \mathbf q = [\mathbf b^\top,\,z,\,\theta_t,\,\mathbf h^\top,\,
    \mathbf a_L^\top,\,\mathbf g_L^\top,\,
  \mathbf a_R^\top,\,\mathbf g_R^\top,\,z_a]^\top,
  \label{eq:wbc_state}
\end{equation}
where $\mathbf b=(x_b,y_b,\psi_b)$ describes the planar base, $z$ the lift height, $\theta_t$ the torso pitch, $\mathbf h\in\mathbb R^2$ the head yaw and pitch, and $\mathbf a_L,\mathbf a_R\in\mathbb R^7$ the arms. The finger coordinates $\mathbf g_L,\mathbf g_R\in\mathbb R^2$ and auxiliary lift-stage coordinate $z_a$ complete the kinematic model, giving $\mathbf q\in\mathbb R^{26}$. Whole-body control acts on the $21$ body coordinates; the two gripper apertures are controlled separately, for a total of $23$ independently commanded DoFs. Measured finger configurations enter the self-collision geometry.

\subsubsection{Composition of local motion policies}
A task map $\mathbf x_i=\phi_i(\mathbf q)$ expresses a behavior in its natural coordinates, with Jacobian $\mathbf J_i=\partial\phi_i/\partial\mathbf q$ and velocity $\dot{\mathbf x}_i=\mathbf J_i\dot{\mathbf q}$. We represent the local policy by a positive-semidefinite metric $\mathbf M_i$ and a generalized force $\mathbf f_i$, satisfying
\begin{equation}
  \mathbf M_i\ddot{\mathbf x}_i+\mathbf f_i=\mathbf0,
  \qquad \mathbf f_i=-\mathbf M_i\mathbf a_i,
  \label{eq:wbc_local_policy}
\end{equation}
where $\mathbf a_i$ is the desired task acceleration. Here, metrics and generalized forces encode motion preferences. We distinguish geometric terms $\mathbf f_i^g$, whose acceleration policies scale quadratically with velocity, from forcing terms $\mathbf f_i^p$ that provide target attraction and damping. Pulling these terms back to configuration space and summing their contributions yields
\begin{equation}
  \begin{aligned}
    \mathbf M &= \sum_i \mathbf J_i^\top\mathbf M_i\mathbf J_i,\\
    \mathbf f^g &= \sum_i \mathbf J_i^\top
    (\mathbf f_i^g+\mathbf M_i\dot{\mathbf J}_i\dot{\mathbf q}),\\
    \mathbf f^p &= \sum_i \mathbf J_i^\top\mathbf f_i^p.
  \end{aligned}
  \label{eq:wbc_pullback}
\end{equation}
The aggregate metric includes contributions from both geometric and forcing policies, while $\dot{\mathbf J}_i\dot{\mathbf q}$ accounts for task-space curvature. We approximate frame and collision-point Jacobian derivatives by directional finite differences along $\mathbf q+\epsilon_J\dot{\mathbf q}$, with $\epsilon_J=10^{-5}$. Restricting the pullback to the enabled coordinates determines which parts of the robot can participate in a task.

\subsubsection{Hand and head pose tracking}
To track translation and orientation in a common Euclidean task space, we represent each frame by seven keypoints: its origin and symmetric offsets along its three axes. For a pose $T=(R,\mathbf p)$, define
\begin{equation}
  \begin{aligned}
    \mathcal R_\delta&=\{\mathbf0,\,\pm\delta\mathbf e_x,
    \pm\delta\mathbf e_y,\,\pm\delta\mathbf e_z\},\\
    \kappa_\delta(T)&=\operatorname{col}_{\mathbf r\in\mathcal R_\delta}
    (\mathbf p+R\mathbf r)\in\mathbb R^{21}.
  \end{aligned}
  \label{eq:wbc_keypoints}
\end{equation}
The corresponding task and target are $\mathbf x=\kappa_\delta(T(\mathbf q))$ and $\mathbf x^\star=\kappa_\delta(T^\star)$. The offset $\delta$ controls the relative influence of orientation error. With $\mathbf e=\mathbf x^\star-\mathbf x$ and $\rho=\|\mathbf e\|$, we define the smooth gating function
\begin{equation}
  G(\rho;r,s)=\tfrac12[1+\tanh(-s(\rho-r))].
  \label{eq:wbc_gate}
\end{equation}
Each pose is tracked by a saturating attraction term with damping that increases near the target:
\begin{equation}
  \begin{aligned}
    \mathbf M_{\rm att}&=m\mathbf I,\\
    \mathbf a_{\rm att}^p&=k\tanh(\alpha\rho)
    \frac{\mathbf e}{\rho+\epsilon}
    -d\,G(\rho;r_d,s_d)\dot{\mathbf x},\\
    \mathbf f_{\rm att}^p&=-\mathbf M_{\rm att}\mathbf a_{\rm att}^p,
  \end{aligned}
  \label{eq:wbc_attractor}
\end{equation}
where $\epsilon=10^{-8}$ regularizes the error direction. The metric weight $m$, attraction gain $k$, damping gain $d$, and damping radius $r_d$ are given in Table~\ref{tab:wbc_attractors}. Both hands follow this forcing policy. Head tracking additionally includes a velocity-dependent geometric attraction with metric $m_H\mathbf I$:
\begin{equation}
  \mathbf a_H^g=\|\dot{\mathbf x}_H\|^2 k_H
  \tanh(\alpha\rho_H)\frac{\mathbf e_H}{\rho_H+\epsilon},
  \qquad \mathbf f_H^g=-m_H\mathbf a_H^g.
  \label{eq:wbc_head_geometry}
\end{equation}
Both head policies contribute to the aggregate metric in Equation~\ref{eq:wbc_pullback}. We use $\delta=0.5\,\mathrm m$ for the hands and $\delta=0.25\,\mathrm m$ for the head. During head-only control, $\delta=5\,\mathrm m$ places greater emphasis on head orientation. The target head height is bounded above by $1.45\,\mathrm m$.

For look-at control, the desired head orientation aligns its forward axis with the direction from the kinematically predicted head position to the look-at point (Appendix~\ref{sec:appendix_quest}). Tracking the resulting pose through the full-body Jacobian coordinates neck orientation with torso posture and lift height.

\begin{table}[t]
  \centering
  \caption{Attractor parameters used in Equation~\ref{eq:wbc_attractor}. All rows use $\alpha=40$ and $s_d=10$. The head row applies separately to its forcing and geometric terms; damping applies only to forcing terms.}
  \label{tab:wbc_attractors}
  \small
  \setlength{\tabcolsep}{4pt}
  \begin{tabular}{lrrrr}
    \toprule
    Task & $m$ & $k$ & $d$ & $r_d$ \\
    \midrule
    Each hand pose & 3 & 60 & 50 & 0.2 \\
    Head pose & 4 & 50 & 100 & 0.05 \\
    Upper-body posture & 0.2 & 50 & 100 & 0.2 \\
    Torso posture & 0.3 & 150 & 100 & 0.2 \\
    Lift posture & 1 & 50 & 5000 & 1.0 \\
    Base yaw attractor & 1 & 50 & 100 & 0.2 \\
    \bottomrule
  \end{tabular}
\end{table}

\subsubsection{Posture regulation and motion allocation}
Posture attraction resolves redundancy by favoring a nominal configuration while the hand and head tasks are tracked. For a selection matrix $\mathbf S$, the task $\mathbf x=\mathbf S\mathbf q$ follows the target $\mathbf S\mathbf q_{\rm nom}$ using Equation~\ref{eq:wbc_attractor}. Separate posture terms govern the torso, lift, and remaining upper-body coordinates, with the weights in Table~\ref{tab:wbc_attractors}. The nominal torso and head angles are zero, and the nominal arm postures are
\begin{equation}
  \begin{aligned}
    \mathbf a_{L,\rm nom}&=(0,1.4,-\pi/2,\pi/2,0,0,0),\\
    \mathbf a_{R,\rm nom}&=(0,1.4,\phantom{-}\pi/2,\pi/2,0,0,0).
  \end{aligned}
  \label{eq:wbc_nominal_arms}
\end{equation}
Leader-specified posture references replace the nominal arm and torso angles and use a nominal lift height of $0.55\,\mathrm m$. Without an external posture reference, the lift target follows its current reference position, so its posture term contributes damping without biasing the height. Posture and pose policies are combined through their metrics, yielding a weighted tradeoff between task tracking and preferred posture.

We further regulate motion allocation with a diagonal metric on the $25$ model coordinates excluding the auxiliary lift stage:
\begin{equation}
  \mathbf M_{\rm pr}=\operatorname{diag}(\mathbf w/\bar w),
  \qquad \mathbf f_{\rm pr}=\mathbf0,
  \qquad \bar w=\frac{1}{25}\sum_{j=1}^{25}w_j.
  \label{eq:wbc_priority}
\end{equation}
We set $w_{\rm lift}=20$, $w_{\rm torso}=10$, and all other entries to $1$. This metric penalizes acceleration of the lift and torso more strongly, encouraging the arms to accommodate target motion while retaining whole-body flexibility.

\subsubsection{Reach-dependent base coordination}
Base translation is weighted according to the reachability of the hand targets. We approximate each arm's workspace by a sphere centered at its shoulder. For the active-hand set $\mathcal A$, the largest target excess beyond these spheres is
\begin{equation}
  d_{\rm reach}=\max_{a\in\mathcal A}
  \bigl(\|\mathbf p_a^\star-\mathbf p_{{\rm shoulder},a}(\mathbf q)\|
  -r_a\bigr),\qquad r_a=0.8\,\mathrm m.
  \label{eq:wbc_reach}
\end{equation}
We take $d_{\rm reach}=-\infty$ when no hand target is active. For the base translation task $\mathbf x_b=(x_b,y_b)$, we use
\begin{equation}
  \begin{aligned}
    \mathbf M_b&=[1+99G(d_{\rm reach};0,100)]\mathbf I_2,\\
    \mathbf f_b^p&=750\dot{\mathbf x}_b.
  \end{aligned}
  \label{eq:wbc_base_translation}
\end{equation}
When all hand targets lie within their workspace spheres, the metric approaches $100\mathbf I_2$, suppressing base motion and favoring arm motion. As a target moves beyond the estimated reach, the metric decreases smoothly toward $\mathbf I_2$, allowing the base to contribute more readily. The constant damping coefficient $750$ moderates translation throughout this transition. The hand and head tasks determine the required base motion through their full-body Jacobians; this term regulates its relative cost.

The desired base heading coordinates the viewing direction with the manipulation targets. Let $\Theta$ contain the planar bearings from the base to the active hand targets and the heading of the desired head frame's forward axis. Their circular mean defines
\begin{equation}
  \psi^\star=\operatorname{atan2}\!\left(
    \sum_{\theta\in\Theta}\sin\theta,
  \sum_{\theta\in\Theta}\cos\theta\right),
  \label{eq:wbc_yaw_target}
\end{equation}
with $\psi^\star$ unwrapped to the nearest angle around $\psi_b$. Bearings to targets coincident with the base are excluded. For an empty $\Theta$, we retain the current heading; when the circular mean is degenerate, we use the arithmetic mean. A scalar attractor with the parameters in Table~\ref{tab:wbc_attractors} tracks $\psi^\star$. A complementary metric-only term regulates yaw motion:
\begin{equation}
  \begin{aligned}
    M_{\psi,\rm reg}&=0.1+49.9G(|e_\psi|;0.09,100),\\
    e_\psi&=\operatorname{wrap}(\psi^\star-\psi_b).
  \end{aligned}
  \label{eq:wbc_yaw_metric}
\end{equation}
The yaw metric increases near the desired heading, reducing unnecessary rotation, and decreases as the heading error grows. This term adds no damping; yaw damping is supplied by the attractor and the configuration-space damping defined below.

\subsubsection{Joint-limit avoidance}
For each bounded coordinate, we define distances to the lower and upper limits as $s^-=q_j-q_j^{\min}$ and $s^+=q_j^{\max}-q_j$. Positive $s$ denotes a configuration within the limit, and $\dot s<0$ indicates motion toward the boundary. Each limit induces the repulsive policy
\begin{equation}
  \begin{aligned}
    \gamma&=\mathbf1[\dot s<0\ \lor\ s<0],\\
    M_{\rm lim}&=\gamma\frac{\eta}
    {\max(s-s_0,\sqrt{\eta/M_{\max}})^2},\\
    a_{\rm lim}&=k_{\rm lim}-\mathbf1[\dot s\leq0]d_{\rm lim}\dot s,
    \qquad f_{\rm lim}=-M_{\rm lim}a_{\rm lim}.
  \end{aligned}
  \label{eq:wbc_joint_limits}
\end{equation}
The activation $\gamma$ engages repulsion when a joint approaches or exceeds its limit. Its metric grows near the boundary and saturates at $M_{\max}$, while the damping component opposes further approach. For the lift, we use $\eta=0.02$, $s_0=0.01$, $M_{\max}=200$, $k_{\rm lim}=3$, and $d_{\rm lim}=8$. For the remaining non-base coordinates, $s_0=0.1$, $M_{\max}=100$, and $k_{\rm lim}=8$, with the same $\eta$ and $d_{\rm lim}$. The bounds $q_j^{\min}$ and $q_j^{\max}$ are specified by the robot model. Commanded arm, torso, head, and lift positions are additionally saturated at these bounds.

\subsubsection{Self-collision avoidance}
We approximate the robot geometry with link-attached spheres and consider a prescribed set of link pairs. For each pair, let $d_i$ be the signed surface separation and $\mathbf x_i,\mathbf p_i$ the closest points on the two bodies. The local task is the surface point $\mathbf x_i$; its Jacobian is evaluated with the instantaneous surface offset fixed in the link frame. We define the separation direction and collision metrics as
\begin{equation}
  \begin{aligned}
    \mathbf n_i&=(\mathbf p_i-\mathbf x_i)/\|\mathbf p_i-\mathbf x_i\|,\\
    \gamma_i&=\mathbf1[d_i\leq d_{\rm engage}\ \land\
    \|\mathbf p_i-\mathbf x_i\|\geq\epsilon],\\
    \mathbf M_i^g=\mathbf M_i^p
    &=\frac{\gamma_i\,0.01}{(1+\epsilon)\max(d_i,0.01)^2}
    \mathbf n_i\mathbf n_i^\top.
  \end{aligned}
  \label{eq:wbc_collision_metric}
\end{equation}
The rank-one metrics concentrate repulsion along the separation direction and grow as the surfaces approach. The activation $\gamma_i$ restricts this influence to nearby surfaces with a defined separation direction. Let $\mathbf c_i=-\mathbf n_i$ for separated surfaces and $\mathbf c_i=\mathbf n_i$ under penetration. The geometric and forcing policies are
\begin{equation}
  \begin{aligned}
    \mathbf a_i^g&=\|\dot{\mathbf x}_i\|^2\mathbf c_i,
    &\mathbf f_i^g&=-\mathbf M_i^g\mathbf a_i^g,\\
    \mathbf a_i^p&=a_{\rm rep}\mathbf c_i,
    &\mathbf f_i^p&=-\mathbf M_i^p\mathbf a_i^p.
  \end{aligned}
  \label{eq:wbc_collision_force}
\end{equation}
We use $d_{\rm engage}=0.15\,\mathrm m$ and $a_{\rm rep}=6$, with no additional local damping. The geometric term scales with point velocity, while the forcing term also produces repulsion at rest. Pullback through the corresponding link Jacobians distributes these avoidance motions across the body. The finite metrics provide soft repulsion rather than hard collision constraints.

\subsubsection{Energization and reference generation}
Energization shapes the geometric motion with respect to a chosen kinetic energy. We combine Euclidean energies in configuration space and head-keypoint space, each with unit weight. Let $\mathbf S_E$ select the same $25$ coordinates as the motion-allocation metric. The energy and its configuration-space representation are
\begin{equation}
  \begin{aligned}
    E&=\tfrac12\|\mathbf S_E\dot{\mathbf q}\|^2
    +\tfrac12\|\mathbf J_H\dot{\mathbf q}\|^2,\\
    \mathbf M_E&=\mathbf S_E^\top\mathbf S_E+\mathbf J_H^\top\mathbf J_H,
    \qquad \mathbf f_E=\mathbf J_H^\top\dot{\mathbf J}_H\dot{\mathbf q}.
  \end{aligned}
  \label{eq:wbc_energy}
\end{equation}
The head contribution to this energy is retained independently of head-target attraction. Both energy and policy pullbacks are restricted to the enabled coordinates. Let $\mathbf M^\dagger$ denote the pseudoinverse of the symmetrized aggregate metric. Energizing the geometric component and adding the forcing terms and configuration-space damping gives
\begin{equation}
  \begin{aligned}
    \mathbf a_g&=-\mathbf M^\dagger\mathbf f^g,\\
    \alpha_E&=-\frac{\dot{\mathbf q}^\top(\mathbf M_E\mathbf a_g+\mathbf f_E)}
    {\dot{\mathbf q}^\top\mathbf M_E\dot{\mathbf q}+10^{-6}},\\
    \mathbf a_0&=-\mathbf M^\dagger
    (\mathbf f^g-\alpha_E\mathbf M\dot{\mathbf q}
    +\mathbf f^p+10\mathbf M\dot{\mathbf q}).
  \end{aligned}
  \label{eq:wbc_energization}
\end{equation}
The correction $\alpha_E$ acts on the geometric component, while target attraction and damping remain forcing terms. We use a configuration-space damping gain of $10$ and a relative pseudoinverse cutoff of $10^{-8}$.

The controller runs at $200\,\mathrm{Hz}$ with $\Delta t=0.005\,\mathrm s$. Before integration, a common scale factor bounds the acceleration while preserving its direction:
\begin{equation}
  \begin{aligned}
    A_j&=\min(a_j^{\max},j_j^{\max}\Delta t/2),\\
    \lambda&=\min\!\left(1,\min_{j:|a_{0,j}|>0}
    \frac{A_j}{|a_{0,j}|}\right),
    \qquad \mathbf a=\lambda\mathbf a_0.
  \end{aligned}
  \label{eq:wbc_acceleration_limit}
\end{equation}
We use $a_j^{\max}=200$ and $j_j^{\max}=7.5\times10^5$ in each coordinate's units. The parameter $j_j^{\max}$ enters through the timestep-dependent acceleration envelope $A_j$; it does not bound the change from the preceding acceleration. At the chosen timestep, $a_j^{\max}$ determines the active bound. Constant-acceleration integration then gives
\begin{equation}
  \begin{aligned}
    \mathbf q_{t+1}&=\mathbf q_t+\Delta t\dot{\mathbf q}_t
    +\tfrac12\Delta t^2\mathbf a_t,\\
    \dot{\mathbf q}_{t+1}&=\dot{\mathbf q}_t+\Delta t\mathbf a_t.
  \end{aligned}
  \label{eq:wbc_integration}
\end{equation}

The integrated trajectory supplies joint-position references for the arms, torso, head, and lift, together with planar velocity references for the base. Upper-body references evolve according to Equation~\ref{eq:wbc_integration}, while the base pose is updated from odometry and its velocity reference is expressed in the measured base frame. Low-level joint and base controllers track these kinematic references.
\par
\endgroup

\subsection{Leader Arm System Details}
\label{sec:appendix_leader}

\textbf{Hardware design.}
The 65\%-scale leader uses 3D-printed links and 18 Dynamixel servos. Six XM430-W210-T servos provide the paired torso drive and actuate the first two joints of each arm; twelve XC330-T288-T servos actuate the remaining five joints per arm and the two gripper triggers. The torso and proximal arm joints support more downstream mass over longer lever arms, producing larger gravitational moments. We select larger motors at these joints to reduce the risk of overheating and stalling, and use two motors at the single torso pivot to share the load. Lower torque demands at the distal joints and triggers allow smaller motors, keeping the interface compact and lightweight. The powered servos operate in torque-control mode through motor-current commands, enabling gravity compensation and null-space posture regularization.

\textbf{Gravity compensation.}
We use Pinocchio's recursive Newton--Euler algorithm with the leader's rigid-body model, including its link masses and centers of mass. Evaluating inverse dynamics at the measured configuration $q$ with zero joint velocities and accelerations isolates the gravitational torques $g(q)$. The backward recursion accumulates the gravitational forces and moments of downstream links to obtain the torque required at each joint. Joint-wise gains set the assistance level through $\tau_g=K_g g(q)$, and the torso support torque is divided equally between its two motors.

\textbf{Null-space posture regularization.}
Each arm is biased toward a comfortable nominal configuration using diagonal stiffness and damping gains $K_p$ and $K_d$. The resulting posture torque is projected into its kinematic null space:
\begin{equation}
  \tau_{\mathrm{post},a}
  =\left(I-J_a^{\dagger}J_a\right)
  \left[K_p(q_a^\star-q_a)-K_d\dot q_a\right].
  \label{eq:appendix_leader_posture}
\end{equation}
Here, $q_a\in\mathbb{R}^7$ contains one arm's joint angles, $q_a^\star$ is its nominal posture, $J_a\in\mathbb{R}^{6\times7}$ is the arm's end-effector Jacobian, and $J_a^{\dagger}$ denotes the Moore--Penrose pseudoinverse. This term encourages comfortable elbow and shoulder configurations while limiting interference with the operator's hand motion.

\textbf{Friction compensation.}
When a joint's measured speed falls below a small threshold, we apply a low-amplitude torque that alternates in sign on successive control cycles. This dithering helps overcome static friction when the operator initiates motion. The dither is disabled once the joint speed exceeds the threshold.

\textbf{Mapping to robot commands.}
At initialization, the operator aligns the leader with a reference posture before teleoperation begins. During teleoperation, we insert the measured torso and arm angles into the full-size robot's kinematic model and use forward kinematics to generate gripper pose attractors for the whole-body controller. We also use the joint angles as nominal robot posture targets to preserve torso and elbow intent. Trigger positions provide continuous gripper commands, while handle buttons and joysticks control lift motion and base velocity, respectively.
\par
\endgroup

\begingroup
\color{black}
\subsection{Quest Teleoperation System Details}
\label{sec:appendix_quest}

\textbf{Unity application.}
The Unity OpenXR application provides the laptop with the headset frame $H$ and the left and right controller frames $C_L,C_R$, each expressed in the common VR world frame $V$. These poses are transmitted through the ROS-TCP bridge in the forward--left--up coordinate convention, together with thumbstick, trigger, and button inputs. We write $T_X^F$ for the pose (position and orientation) of frame $X$ expressed in reference frame $F$; it transforms coordinates from $X$ to $F$. Thus, the laptop receives $T_H^V,T_{C_L}^V,T_{C_R}^V$.

In this notation, $R_X^F$ is the rotation part of $T_X^F$, and $z_X^F$ is the height of frame $X$'s origin along the vertical axis of $F$. An additional subscript $0$ denotes the value recorded at calibration, as in $T_{X,0}^F$ and $z_{X,0}^F$; time indices on live poses are omitted. A superscript $*$ denotes a desired target value.

\textbf{Calibration.}
Calibration establishes a consistent upright human hip reference corresponding to the robot's hip frame and compensates for the headset's tilt when worn around the neck. With the operator standing upright, we project the headset's forward axis onto the horizontal plane and align its vertical axis with world up, obtaining the level headset orientation $R_{\mathrm{lev},0}^{V}$. We store the local orientation correction
\begin{equation}
  Q=(R_{H,0}^{V})^{\top}R_{\mathrm{lev},0}^{V}.
  \label{eq:appendix_quest_calibration}
\end{equation}
A fixed headset-to-hip offset applied to this level headset pose estimates the initial human hip frame $U$. We record $z_{U,0}^{V}$, the human hip height in the VR world at calibration, and $z_{\mathrm{head},0}^{W}$, the robot head height in robot world frame $W$ at calibration. The two world origins need not coincide.

\textbf{Retargeting.}
During operation, we define a corrected headset frame $\widetilde H$ by applying $R_{\widetilde H}^{V}=R_H^VQ$ while retaining the measured position. This removes the initial wearing tilt while preserving subsequent body lean. The fixed transform $T_U^{\widetilde H}$ locates the estimated hip relative to the corrected headset, giving the live hip pose $T_U^V=T_{\widetilde H}^{V}T_U^{\widetilde H}$. We then construct a level reference $\bar U$ whose horizontal position follows $U$ and whose heading points from the estimated hip toward a point in front of the corrected headset. Its height remains at the calibration value $z_{U,0}^{V}$, preserving vertical hand motion during crouching.

Each controller pose is expressed relative to $\bar U$ and transferred to the robot hip $R$, located on the lift below the torso pivot:
\begin{equation}
  T_{G_i}^{W}
  =\operatorname{Trans}_{W}(\delta_z\mathbf e_z)\,
  T_R^W\left(T_{\bar U}^{V}\right)^{-1}T_{C_i}^{V}.
  \label{eq:appendix_quest_hands}
\end{equation}
Here, $G_i$ is the gripper target frame corresponding to controller $C_i$, with $i\in\{L,R\}$. The operator $\operatorname{Trans}_W(\mathbf d)$ applies a pure translation by $\mathbf d$ expressed in $W$, and $\mathbf e_z=[0,0,1]^\top$ is the unit vertical vector. The robot hip pose $T_R^W$ comes from its kinematic model. The mapping transfers relative hand position and orientation without rescaling; the adjustable vertical-workspace offset $\delta_z$ shifts both hand and head targets.

For torso and head motion, let $\mathbf a=(R_{\bar U}^{V})^{\top}R_{\widetilde H}^{V}\mathbf e_z$ be the corrected headset's vertical axis in the live heading frame. We compute the lean angle $\alpha$ from its forward and vertical components $a_x,a_z$, and the desired head height $z_{\mathrm{head}}^*$ in $W$ from the change in hip height:
\begin{equation}
  \begin{aligned}
    \alpha &= \operatorname{atan2}(a_x,a_z),\\
    z_{\mathrm{head}}^{*} &= z_{\mathrm{head},0}^{W}
    +\left(z_U^V-z_{U,0}^{V}\right)+\delta_z.
  \end{aligned}
  \label{eq:appendix_quest_head}
\end{equation}
Measuring lean in the heading frame preserves its sign as the operator turns. Let $\mathbf q_t$ denote the robot configuration at the current time $t$, with torso-angle component $q_{\mathrm{torso}}$. A scaled and bounded version of $\alpha$ defines the candidate torso angle $\theta_{\mathrm{lean}}$. Replacing $q_{\mathrm{torso}}$ with this angle in forward kinematics gives the desired head coordinates $(x_{\mathrm{head}}^*,y_{\mathrm{head}}^*)$ in $W$. This shapes the attractor; the whole-body controller determines the resulting joint motion.

\textbf{Head orientation computation.}
Let $\mathbf p_{G_i}^{W}$ denote the position of gripper attractor $G_i$, and $\widehat{\mathbf p}_{\mathrm{head}}^{W}$ the robot head position returned by the forward kinematics used above. The default look-at point is
\begin{equation}
  \mathbf p_{\mathrm{look}}^{W}
  =\frac{\mathbf p_{G_L}^{W}+\mathbf p_{G_R}^{W}}{2}
  +h_{\mathrm{look}}\mathbf e_z,
  \label{eq:appendix_quest_look_point}
\end{equation}
where $h_{\mathrm{look}}=0.20\,\mathrm m$ is the configured vertical offset above the manipulation location, as in Equation~\ref{eq:main_head_gaze}. Selecting either hand replaces the midpoint with that gripper's position. We construct the desired head axes, all expressed in $W$, as
\begin{equation}
  \begin{aligned}
    \mathbf x^* &=\frac{\mathbf p_{\mathrm{look}}^{W}-\widehat{\mathbf p}_{\mathrm{head}}^{W}}
    {\|\mathbf p_{\mathrm{look}}^{W}-\widehat{\mathbf p}_{\mathrm{head}}^{W}\|_2},\\
    \widetilde{\mathbf z} &=\mathbf e_z-(\mathbf e_z^\top\mathbf x^*)\mathbf x^*,\\
    \mathbf z^* &=\frac{\widetilde{\mathbf z}}{\|\widetilde{\mathbf z}\|_2},
    \qquad \mathbf y^*=\mathbf z^*\times\mathbf x^*,\\
    R_{\mathrm{head}}^{W,*} &=[\,\mathbf x^*\;\mathbf y^*\;\mathbf z^*\,].
  \end{aligned}
  \label{eq:appendix_quest_head_orientation}
\end{equation}
Here, $\mathbf x^*,\mathbf y^*,\mathbf z^*$ are the forward, left, and up axes, and $R_{\mathrm{head}}^{W,*}$ is the desired head rotation in robot world coordinates. The forward axis points toward the look-at point; projecting world up onto its perpendicular plane fixes the remaining rotation about that axis. For a nearly vertical look direction, we project the model-predicted head up axis instead, using world $y$ if that projection also degenerates. A negligible look-direction length produces no new head target. This determines gaze independently of the neck-mounted headset's orientation.

\textbf{Base coordination and clutching.}
In VR Joystick mode, thumbsticks command planar base velocity. We integrate this command one control interval forward before computing the robot hip and head transforms, so the arm and head targets advance with base motion instead of lagging behind it. In VR Body Tracking mode, engaging the base clutch anchors the human hip reference and robot base pose. Subsequent human translation and yaw define the base reference used to construct the attractors, with translation scaled and expressed in the anchored robot heading. The whole-body controller coordinates base motion to track these targets. Releasing the base clutch blends back to joystick control; re-engaging resets the anchor. A separate teleoperation clutch holds the upper-body attractors relative to the robot hip when released. Triggers provide continuous gripper-position commands.
\par
\endgroup

\begingroup
\color{black}
\subsection{Sensorized Human Data System Frame Calibration}
\label{sec:appendix_human_calibration}

\textbf{Purpose and frame convention.}
Frame calibration aligns the controller and Quest headset tracking frames on the human system with the gripper and head tracking frames on the robot. We write $T_X^F$ for the pose of frame $X$ expressed in frame $F$, mapping coordinates from $X$ to $F$, and use $V$ for the Quest world frame.

\textbf{Intrinsic and extrinsic calibration.}
We estimate camera intrinsics and lens distortion from multiple views of an AprilGrid target. For extrinsic calibration, the target remains stationary while the camera--tracker assembly moves through different positions and orientations. Images are paired with time-stamped tracker poses, and perspective-$n$-point estimation recovers the target pose in the camera frame. An OpenCV hand--eye solve using the Park method then estimates the fixed camera-to-tracker transform. We apply the same procedure on the robot, replacing Quest tracking with forward kinematics from measured joint configurations and keeping the mobile base stationary.

\textbf{Alignment through the camera frame.}
For each assembly, let $S$ denote the human-side tracking frame: a controller frame $C_i$ for hand $i\in\{L,R\}$, or the Quest headset frame $H$. Let $A$ denote the corresponding robot gripper tracking frame $G_i$ or head tracking frame $\mathrm{head}$, and $K$ the associated camera frame. The matched camera placement and coordinate conventions provide a shared reference between the two systems. The wearable calibration gives $T_K^S$, while the robot calibration gives $T_K^A$. These transforms define the camera-based alignment of the measured pose $T_S^V$:
\begin{equation}
  T_A^V = T_S^V\,T_K^S\left(T_K^A\right)^{-1}.
  \label{eq:appendix_human_frame_alignment}
\end{equation}
Applying this mapping to the tracked poses produces gripper and head trajectories in the robot's task-space convention for learning and control.
\par
\endgroup

\begingroup
\color{black}
\subsection{Data Collection and Pipelining Details}
\label{sec:appendix_data_pipeline}

\textbf{Data collection and storage.}
Robot and sensorized human demonstrations are recorded as asynchronous ROS streams, including camera images, tracked poses, gripper states, and, on the robot, joint states and control targets. For recordings spanning multiple computers, system clocks are synchronized using chrony with the laptop as the local time reference. Each sample retains its message-header timestamp and its reception timestamp, allowing source timing and reception delay to be distinguished. Episodes are stored in HDF5 with separate datasets for each stream, retaining compressed image payloads alongside numerical data and recording metadata. A separate writer process batches queued samples, keeping disk writes outside the subscriber callbacks.

\textbf{Offline temporal synchronization.}
Before training, we restrict each episode to the interval shared by its recorded streams: from the latest stream start to the earliest stream end. We construct a uniform grid with period $\Delta=1/f$, where $f=50\,\mathrm{Hz}$, starting at the beginning of this overlap. Let $g_t$ denote the grid time at step $t$, and let $\tau_{i,j}$ and $y_{i,j}$ denote the source timestamp and value of sample $j$ in stream $i$. For samples ordered by timestamp, each synchronized value is taken from the most recent sample at or before the grid time:
\begin{equation}
  j_i(t)=\max\bigl\{j:\tau_{i,j}\leq g_t\bigr\},
  \qquad \widetilde{y}_{i,t}=y_{i,j_i(t)}.
  \label{eq:appendix_data_sync}
\end{equation}
This preceding-sample rule applies to both images and numerical signals: no sample later than a grid time is used at that step. Samples may be repeated or omitted without temporal interpolation. We use the same rule during deployment to minimize differences in temporal sampling between training and inference. The synchronized streams share grid timestamps and sequence lengths in the output HDF5 dataset.

\textbf{Training sample construction.}
At synchronized timestep $t$, the observation $\mathbf{o}_t$ contains RGB images from the head and two wrist cameras. It is paired with a language instruction $\ell_t$ and future action targets $\mathbf{A}_t=[\mathbf{a}_{t+1},\ldots,\mathbf{a}_{t+H}]$, where $H$ is the prediction horizon. Targets extending beyond an episode are padded and excluded from the training loss. Robot-specific proprioception is omitted from the policy input. Saved camera specifications, temporal sampling settings, action definitions, and normalization statistics preserve the training data interface during deployment.

\textbf{Online observation synchronization.}
At inference, incoming camera samples are retained in timestamped buffers. Let $\tau_i^{\mathrm{latest}}$ be the most recent source timestamp currently buffered for required stream $i$. Once all required streams are available, the observation reference time is
\begin{equation}
  g_* = \Delta\left\lfloor
  \frac{\min_i\tau_i^{\mathrm{latest}}}{\Delta}
  \right\rfloor.
  \label{eq:appendix_online_data_sync}
\end{equation}
The latest buffered sample at or before $g_*$ is selected from each camera, using the same preceding-sample rule as offline synchronization. Thus, offline processing traverses an entire shared recording interval, whereas online processing assembles an observation at the latest grid time supported by every required stream (Figure~\ref{fig:appendix_data_sync}). Excessive sample-to-grid gaps generate diagnostic warnings.

\begin{figure}[t]
  \centering
  \includegraphics[width=\columnwidth]{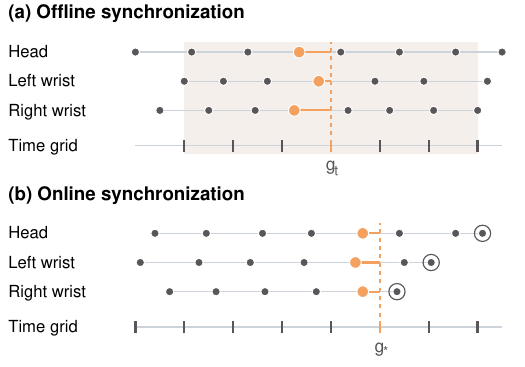}
  \caption{\textbf{Offline and online temporal synchronization.} Dots indicate source samples; orange dots show the latest samples at or before the highlighted grid time. (a) Offline synchronization resamples the shared recording interval (shaded), with one grid time illustrated. (b) Online synchronization selects from currently buffered samples. Rings mark the latest sample in each stream; the earliest of these determines the reference time after rounding down to the grid. Timing is schematic.}
  \label{fig:appendix_data_sync}
\end{figure}
\par
\endgroup

\begingroup
\color{black}
\subsection{Policy Deployment Details}
\label{sec:appendix_policy_deployment}

\textbf{Asynchronous inference and execution.}
Policy inference runs in a background worker while a separate execution loop advances through queued actions at the policy data rate. Each request uses synchronized observations and the current language instruction, with the action reference frame captured at launch. New requests are launched after a configured number of action steps, with one prediction evaluated at a time. During inference, the robot continues executing the previous chunk; an independent command loop supplies the current targets to the whole-body controller. If the queue is exhausted, the runtime retains a hold target until a new chunk becomes available. This separates variable inference latency from periodic command delivery.

\textbf{Real-time chunking.}
We use training-time action conditioning for real-time chunking (RTC)~\cite{black2025training}. Let $H$ denote the action horizon and $d<H$ the estimated inference delay in policy steps. The next $d$ queued actions form a committed prefix: these actions are expected to execute while the next prediction is computed. The flow sampler fixes this prefix throughout its Euler integration and generates the remaining actions conditioned on it. During training, randomly sampled prefix lengths expose the model to this conditioning, with supervision applied only to the remaining valid actions. The runtime estimates $d$ from recent handoff delays, bounded by the delay range used in training; the initial prediction uses no prefix. When a prediction returns, the runtime discards the committed prefix and any additional steps already executed during inference before replacing the queue with the remaining actions.

\textbf{Reference-frame consistency.}
The local RTC implementation constructs its prefix from queued world-frame targets and re-expresses them in the new chunk's relative-base frame. Let $B_t$ be the gravity-aligned pseudo-base frame derived from the most recent commanded head pose at request step $t$, initialized from the measured pose when no prior target exists. For a queued target of the head or hand $j\in\{\mathrm{head},L,R\}$ at future step $t+k$, its prefix pose is
\begin{equation}
  T_{j,t+k}^{B_t}
  =\left(T_{B_t}^{W}\right)^{-1}T_{j,t+k}^{W},
  \label{eq:appendix_rtc_frame_alignment}
\end{equation}
where $W$ is the world frame, $T_{B_t}^{W}$ is the pseudo-base pose, and $T_{j,t+k}^{W}$ is the queued world-frame target. The resulting prefix is normalized using the training statistics. The same reference-frame snapshot converts the returned relative predictions into world-frame attractors, preserving their interpretation as the robot moves during inference.

\textbf{Local and remote inference.}
The reported policy evaluations use local inference on the robot's GPU laptop, typically using an NVIDIA RTX~5090 GPU. Alternatively, a remote server executes the policy, receiving preprocessed camera arrays and the language instruction over a WebSocket connection and returning a predicted action chunk. The server supplies the checkpoint's input/output specifications and normalization statistics when the connection is established. Observation assembly, action decoding, queue management, and whole-body control remain on the robot in both modes. For RTC, the local backend supplies the explicitly re-expressed prefix above; the remote implementation retrieves its prefix from cached predictions using the execution index and delay supplied by the client.

\par
\endgroup

\begingroup
\color{black}
\subsection{High-Level Policy Details}
\label{sec:appendix_high_level_policy}

\textbf{Models and inference configuration.}
The high-level policy uses Gemini~3.5 Flash (\texttt{gemini-3.5-flash}) through the Gemini API for two operations: translating user requests into skill sequences and assessing execution from head-camera observations. Each operation has a separate system prompt and JSON response schema. The planner returns a task, informational answer, or unsupported-request decision; the monitor returns an execution status and supporting evidence. A local controller validates these responses and manages policy execution. Table~\ref{tab:appendix_high_level_vlm} summarizes the inference settings. Automatic context caching attempts to cache reusable system prompts of at least 4096 tokens for one hour.

\begin{table}[t]
  \centering
  \caption{High-level VLM configuration.}
  \label{tab:appendix_high_level_vlm}
  {\small
    \setlength{\tabcolsep}{3pt}
    \renewcommand{\arraystretch}{1.08}
    \begin{tabular}{@{}p{0.44\columnwidth}p{0.51\columnwidth}@{}}
      \toprule
      \textbf{Parameter} & \textbf{Setting} \\
      \midrule
      Temperature & 0.1 \\
      Maximum output tokens & 4096 \\
      Image resolution / API tier & Low / priority \\
      Response format & Schema-constrained JSON \\
      Thinking level & Planning: low; monitoring: minimal \\
      Request timeout & Planning: 8\,s; monitoring: 4\,s \\
      Nominal monitoring interval & 6\,s \\
      API attempts & Up to 3; retry delays of 1 and 2\,s, subject to timeout \\
      \bottomrule
    \end{tabular}
  }
\end{table}

\textbf{Skill library and task planning.}
Each supported skill has a semantic description, a fixed language instruction for the low-level policy, and visual completion criteria (Table~\ref{tab:appendix_high_level_skills}). The planner receives the user request, the configured skill identifiers and descriptions, the unfinished portion of any existing plan, and the latest head-camera image when available. It preserves explicitly requested task order and rejects the entire request if any requested task cannot be mapped to the library. The proposed plan requires user confirmation before execution. Once approved, each selected identifier resolves locally to its fixed policy instruction.

\begin{table*}[t]
  \centering
  \caption{Configured high-level skill library.}
  \label{tab:appendix_high_level_skills}
  {\small
    \setlength{\tabcolsep}{3pt}
    \renewcommand{\arraystretch}{1.08}
    \begin{tabular}{@{}p{0.14\textwidth}p{0.39\textwidth}p{0.43\textwidth}@{}}
      \toprule
      \textbf{Skill} & \textbf{Low-level policy instruction} & \textbf{Visual completion criteria} \\
      \midrule
      Tidy toys & Pick up the toys from the table and place them neatly in the toy basket. & No toys remain on the table, and the toys are visibly in the basket. \\
      Tidy pillows & Navigate to the couch and tidy the pillows. & Both pillows are fully visible, upright, and neatly arranged on the couch. \\
      Hang clothing & Grab the clothing item from the person and hang it on the clothes rack. & The clothing item is visibly hanging on the rack (a wooden pole). \\
      Deliver jacket & Grab the jacket from the clothes rack and give it to the person. & The requesting person visibly receives the jacket. \\
      Clean table & Pick up the handheld vacuum and clean the dining table. & The table is visibly clean after vacuuming, and the vacuum is retracted from the surface. \\
      Return vacuum & Put the vacuum back. & The handheld table vacuum is visibly returned to storage. \\
      Fetch snack & Fetch a snack and bring it to the person. & The snack is delivered to the requesting person. \\
      \bottomrule
    \end{tabular}
  }
\end{table*}

\par\smallskip
\noindent\fbox{%
  \begin{minipage}{\dimexpr\columnwidth-2\fboxsep-2\fboxrule\relax}
    \small
    \textbf{Planning prompt example (abbreviated).}
    Return only a JSON response conforming to the task-decision schema. Select only configured skill IDs. Preserve every supported task in the user's explicit order and require confirmation. If any requested task is unsupported, reject the entire request. Do not author low-level policy instructions.

    \smallskip
    \textit{Example user request:} \lq\lq{}Tidy the couch pillows, then fetch a snack.\rq\rq

    \smallskip
    \textit{Illustrative ordered skill selection:}
    \texttt{tidy\_couch\_pillows}, then \texttt{fetch\_snack}.
\end{minipage}}
\par\smallskip

\textbf{Visual memory and keyframe selection.}
The image histories provide candidate observations at two time scales; only selected views are passed to the VLM. Each monitoring request uses up to four live head-camera images: the pinned step-start observation, an intermediate keyframe from the sparse history nearest the temporal midpoint between step start and the latest observation, a recent image from the short-term buffer approximately 3\,s before the latest observation, and the latest image. The sparse history provides longer-term context for identifying the current stage of the task and tracking overall progress. The recent buffer supports comparison of ongoing motion with the latest view, providing recent visual evidence for judging whether the task is complete or the robot is stuck.

The sparse history retains at most 12 keyframes over 180\,s. After at least 15\,s, a new keyframe is selected when its 64-bit grayscale average hash differs from the previous keyframe by at least 18 bits; after 45\,s, a frame is retained regardless of change. The recent buffer instead samples every 0.5\,s over an 8\,s window, with at most 20 frames.

\textbf{Execution monitoring and completion decisions.}
Each monitoring request includes the current skill's success criteria, the latest six available textual assessments, and the ordered live views. Each configured skill also supplies one completion-reference image, placed before the live observations. This reference illustrates an acceptable outcome. The monitor is instructed to require all success criteria to be visible in the latest observation together with a task-relevant change from the step-start view.

Monitoring runs nominally every 6\,s.

\par\smallskip
\noindent\fbox{%
  \begin{minipage}{\dimexpr\columnwidth-2\fboxsep-2\fboxrule\relax}
    \small
    \textbf{Monitoring prompt example (abbreviated).}
    Assess only the supplied skill. Treat the completion reference as an example of an acceptable outcome. Return completed only if the latest live view satisfies every criterion and shows a relevant change from the start; otherwise return uncertain when visual evidence is insufficient. Do not infer completion from elapsed time.

    \smallskip
    \textit{Skill criteria:} Both couch pillows must be fully visible, upright, and neatly arranged; navigation alone is insufficient.

    \smallskip
    \textit{Image order:} Completion reference; step start; intermediate; recent; latest.
\end{minipage}}
\par\smallskip

\textbf{Speech input and spoken feedback.}
Speech recognition runs locally through NVIDIA NeMo using \texttt{nvidia/parakeet-unified-en-0.6b}, a 600-million-parameter English model with a 24-layer Unified FastConformer encoder and recurrent neural network transducer decoder~\cite{nvidia_parakeet_unified}. Incoming audio is segmented using WebRTC voice activity detection and an amplitude threshold, and each completed utterance is then transcribed. Accepted text enters the same planning interface as typed instructions.

Spoken feedback is synthesized locally by Piper using an American English female voice with 22.05\,kHz output~\cite{piper_hfc_female_voice}. The announcer speaks user-facing notifications marked for speech, including plan confirmation, task completion, and requests for assistance.

\par
\endgroup

\begingroup
\color{black}
\subsection{Low-Level Policy Details}
\label{sec:appendix_low_level_policy}

\textbf{Visual representation.}
The policy receives synchronized head and wrist images without robot-specific proprioception. Each view is encoded by a pretrained DINOv3 ViT-B/16 backbone~\cite{simeoni2025dinov3}. We process head images at $384\times384$ and retain all tokens to preserve distant objects and spatial relationships across the scene. Wrist images are processed at $224\times224$, retaining only the class and register tokens. These close-range views emphasize interactions around the grippers, allowing a smaller cross-attention token budget. Learned camera embeddings identify each view, and the concatenated visual tokens are layer-normalized before language conditioning.

\textbf{Language conditioning.}
A frozen DistilBERT encoder~\cite{sanh2020distilbertdistilledversionbert} maps instruction $\ell_t$ to a 768-dimensional embedding $\mathbf{e}_t$ by mean-pooling non-padding token features. A learned network produces scale and shift vectors for FiLM~\cite{perez2017filmvisualreasoninggeneral}:
\begin{equation}
  \widetilde{\mathbf{z}}_{t,i}
  =\bigl(\mathbf{1}+\boldsymbol{\gamma}(\mathbf{e}_t)\bigr)
  \odot\mathbf{z}_{t,i}+\boldsymbol{\beta}(\mathbf{e}_t),
  \label{eq:appendix_low_level_film}
\end{equation}
where $\mathbf{z}_{t,i}$ is visual token $i$, $\boldsymbol{\gamma}$ and $\boldsymbol{\beta}$ are the predicted scale and shift, and $\odot$ denotes element-wise multiplication. The modulation is shared across visual tokens and initialized to the identity.

\textbf{FlowDiTX action generator.}
FlowDiTX projects a noisy action chunk $\mathbf{X}_s\in\mathbb{R}^{H\times29}$ into 768-dimensional tokens with learned temporal position embeddings, where $H$ is the action horizon and $s\in[0,1]$ is flow time. Each of six blocks applies temporal self-attention, cross-attention, and a feed-forward network. Action tokens supply the cross-attention queries, while language-conditioned visual tokens supply the keys and values. A sinusoidal flow-time embedding controls adaptive layer normalization and initially zero residual gates. The output head predicts an action-space velocity field, which is integrated to generate the action chunk.

\textbf{Action representation.}
We represent rotations by the first two columns of their rotation matrices~\cite{zhou2019continuity}. This continuous 6D representation avoids Euler-angle wrapping and singularities and the sign ambiguity of unit quaternions. The target columns are orthonormal unit vectors with components in $[-1,1]$, so we retain their fixed scale without dataset standardization. Predicted columns are orthonormalized when decoded into rotations. Position and gripper targets are standardized using dataset statistics.

\textbf{Training procedure.}
We train for 150,000--200,000 optimizer updates, jointly fine-tuning the visual backbones with the action generator and FiLM network while keeping DistilBERT frozen. Flow times are sampled from $\mathrm{Beta}(1,1.5)$, emphasizing small $s$ and therefore noisier actions. Following the rationale in $\pi_0$~\cite{BlackK-RSS-25}, we assign more training weight to this regime because robot observations strongly constrain appropriate actions, making observation-conditioned prediction demanding even when the noisy action chunk provides little information. Training-time RTC supplies a randomly selected clean prefix; the flow-matching loss excludes both the prefix and episode padding.

We use AdamW with separate vision and policy learning rates, linear warm-up followed by cosine decay, and gradient-norm clipping (Table~\ref{tab:appendix_low_level_hyperparameters}). Euler integration generates action chunks at inference; execution is described in Appendix~\ref{sec:appendix_policy_deployment}.

\begin{table}[t]
  \centering
  \caption{Low-level policy training hyperparameters.}
  \label{tab:appendix_low_level_hyperparameters}
  {\small
    \setlength{\tabcolsep}{3pt}
    \renewcommand{\arraystretch}{1.06}
    \begin{tabular}{@{}p{0.49\columnwidth}p{0.46\columnwidth}@{}}
      \toprule
      \textbf{Hyperparameter} & \textbf{Value} \\
      \midrule
      Vision backbone & DINOv3 ViT-B/16 \\
      Vision parameters per view & Approximately 86M \\
      Language encoder & Frozen DistilBERT \\
      Language embedding dimension & 768 \\
      Visual context length & 591 tokens \\
      Token / hidden dimension & 768 \\
      Transformer blocks & 6 \\
      Attention heads & 8 \\
      Feed-forward hidden dimension & 3072 \\
      Flow-time embedding dimension & 256 \\
      Action transformer parameters & 90.3M \\
      Action dimension / horizon & 29 / 50 \\
      Context / transformer dropout & 0.1 \\
      FiLM input dropout & 0.2 \\
      Optimizer & AdamW \\
      Policy learning rate & $10^{-4}$ \\
      Vision learning rate & $10^{-5}$ \\
      Global batch size & 512 ($4$ GPUs $\times128$) \\
      Optimizer updates & 150,000--200,000 \\
      Learning-rate schedule & Cosine decay \\
      Linear warm-up & 5,000 updates \\
      AdamW $(\beta_1,\beta_2)$ & $(0.95,0.999)$ \\
      Weight decay / $\epsilon$ & $10^{-6}$ / $10^{-8}$ \\
      Gradient clipping norm & 1.0 \\
      Mixed precision & bfloat16 \\
      Flow-time distribution & $\mathrm{Beta}(1,1.5)$ \\
      RTC prefix length & Uniform over 0--6 steps \\
      Euler integration steps & 10 \\
      \bottomrule
    \end{tabular}
  }
\end{table}

\textbf{Image augmentation.}
Random resized crops and small rotations vary image framing, while color jitter perturbs illumination and color to encourage robustness to visual changes (Table~\ref{tab:appendix_image_augmentation}). Evaluation uses deterministic resizing and the same ImageNet normalization.

\begin{table}[t]
  \centering
  \caption{Image augmentation settings.}
  \label{tab:appendix_image_augmentation}
  {\small
    \setlength{\tabcolsep}{3pt}
    \renewcommand{\arraystretch}{1.06}
    \begin{tabular}{@{}p{0.49\columnwidth}p{0.46\columnwidth}@{}}
      \toprule
      \textbf{Hyperparameter} & \textbf{Value} \\
      \midrule
      Random resized crop area & 0.9--1.0 \\
      Random rotation & $\pm5^\circ$ \\
      Color jitter: brightness & 0.3 \\
      Color jitter: contrast & 0.4 \\
      Color jitter: saturation & 0.5 \\
      Color jitter: hue & 0.08 \\
      RGB normalization mean & $(0.485,0.456,0.406)$ \\
      RGB normalization std. & $(0.229,0.224,0.225)$ \\
      \bottomrule
    \end{tabular}
  }
\end{table}

\par
\endgroup

\begingroup
\color{black}
\subsection{Task Details}
\label{sec:appendix_task_details}

Figure~\ref{fig:appendix_task_sequences} illustrates representative robot executions of the six evaluation tasks. Appendix~\ref{sec:appendix_task_scoring} provides their progress-based scoring rubrics.

\begin{figure*}[p]
  \centering
  \color{black}
  \includegraphics[width=\textwidth,height=0.88\textheight,keepaspectratio]{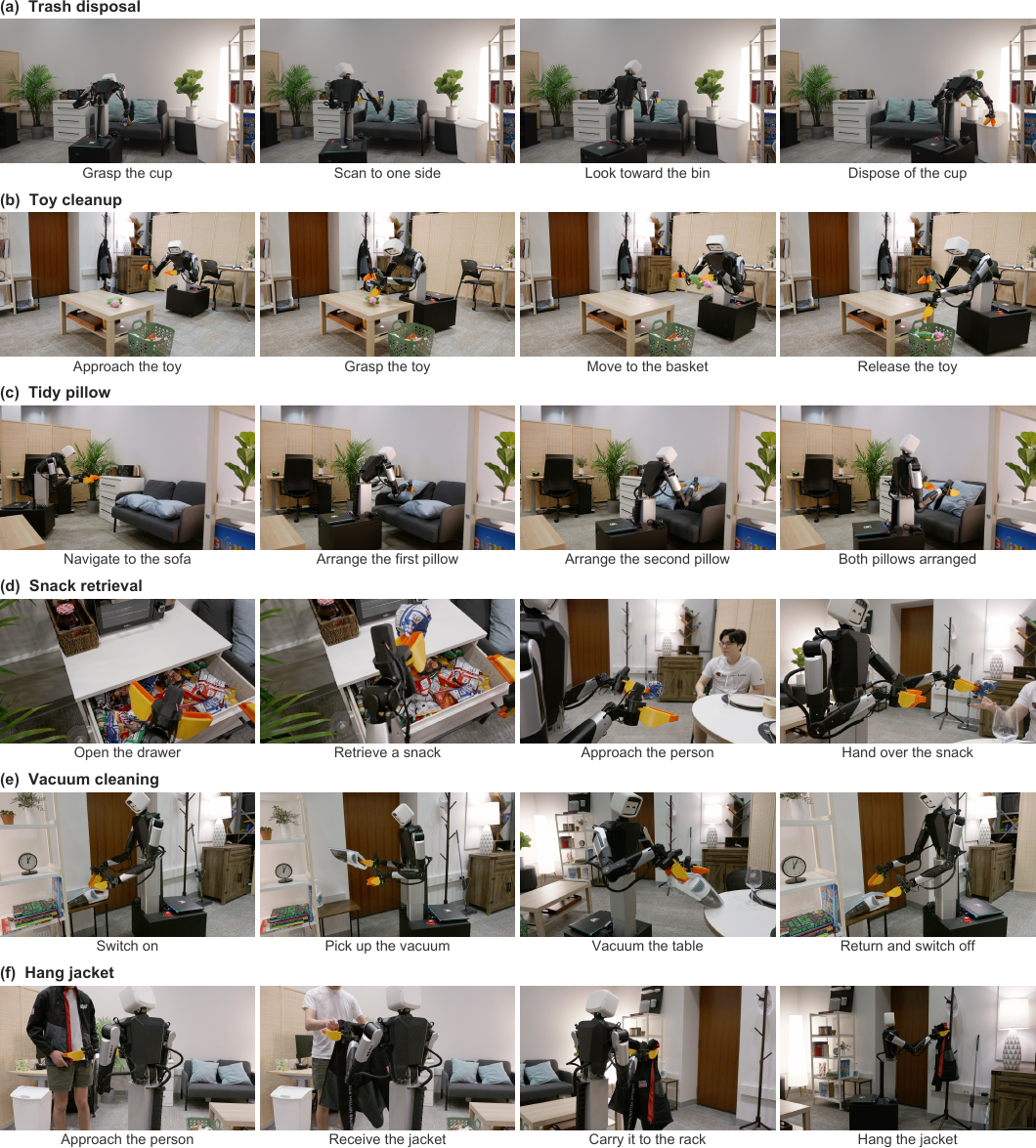}
  \caption{\textbf{Representative robot task sequences.} Each row shows four key stages, ordered from left to right. Rows (a--c) correspond to the sensorized human data tasks, and rows (d--f) to the robot teleoperation data tasks.}
  \label{fig:appendix_task_sequences}
\end{figure*}
\par
\endgroup

\subsection{Progress-Based Task Scoring Rubrics}
\label{sec:appendix_task_scoring}

Each rollout is assigned the cumulative score associated with the furthest task milestone completed. A milestone is credited only when its specified end state is achieved. Rubric scores are normalized to \([0,1]\) and listed to three decimal places, where \(0.000\) denotes no task-relevant progress and \(1.000\) denotes complete task success.

\begin{table}[t]
  \centering
  \caption{Progress-based task scoring rubrics. Each score is cumulative.}
  \label{tab:task_scoring}
  {\small
    \setlength{\tabcolsep}{3pt}
    \renewcommand{\arraystretch}{1.05}
    \begin{tabular}{p{0.20\columnwidth}cp{0.60\columnwidth}}
      \hline
      \textbf{Task} & \textbf{Score} & \textbf{Condition} \\
      \hline

      \textbf{Toy Cleanup}
      & \(0.000\) & No task-relevant progress. \\
      & \(0.500\) & The toy is grasped and lifted from the table. \\
      & \(1.000\) & The toy is released inside the basket. \\
      \hline

      \textbf{Tidy Pillow}
      & \(0.000\) & No task-relevant progress. \\
      & \(0.143\) & The robot navigates to the couch. \\
      & \(0.429\) & The first pillow is picked up. \\
      & \(0.571\) & The first pillow is placed upright against the seat back. \\
      & \(0.857\) & The second pillow is picked up. \\
      & \(1.000\) & The second pillow is placed upright against the seat back. \\
      \hline

      \textbf{Trash Disposal}
      & \(0.000\) & No task-relevant progress. \\
      & \(0.400\) & The cup is picked up from the couch. \\
      & \(0.800\) & The robot reaches the trash can while retaining the cup. \\
      & \(1.000\) & The cup is released inside the trash can. \\
      \hline

      \textbf{Vacuum Cleaning}
      & \(0.000\) & No task-relevant progress. \\
      & \(0.250\) & The vacuum is switched on. \\
      & \(0.375\) & The vacuum is picked up. \\
      & \(0.625\) & The spilled debris is vacuumed from the dining table. \\
      & \(0.750\) & The vacuum is returned to its storage location. \\
      & \(1.000\) & The vacuum is switched off. \\
      \hline

      \textbf{Hang Jacket}
      & \(0.000\) & No task-relevant progress. \\
      & \(0.286\) & The robot navigates to the person. \\
      & \(0.429\) & The robot receives and secures the jacket. \\
      & \(0.571\) & The jacket is transported to the clothes rack. \\
      & \(1.000\) & The jacket is successfully hung on the rack. \\
      \hline

      \textbf{Snack Retrieval}
      & \(0.000\) & No task-relevant progress. \\
      & \(0.167\) & The drawer is opened. \\
      & \(0.500\) & The snack is retrieved from the drawer. \\
      & \(0.667\) & The drawer is closed. \\
      & \(1.000\) & The snack is delivered to the requesting person. \\
      \hline

    \end{tabular}
  }
\end{table}

\begingroup
\color{black}
\subsection{Workspace Analysis Details}
\label{sec:appendix_workspace}

The workspace comparison in Table~\ref{tab:workspace_comparison} measures how closely each embodiment's fixed-base reach corresponds to human reach. We estimate these workspaces by sampling feasible joint ranges, evaluating end-effector positions through forward kinematics, and refining the boundaries of the resulting volumetric representations.

\subsubsection{Kinematic sampling}
For each embodiment, we sample the arm joints together with shared body joints, such as the torso and lift, while holding the mobile base stationary. Shared-joint motion contributes to both arms' reach. Joint ranges and mechanical couplings follow the respective kinematic models. Forward kinematics maps each sampled configuration to left- and right-endpoint positions, which are accumulated into separate workspaces. All models retain their physical dimensions and share a ground plane and forward direction, using the relative placement described in the main paper.

We use scrambled Sobol sequences to cover the joint-limit domain. Additional samples concentrate near joint limits and on selected boundary faces, improving coverage of configurations that produce extreme reaches. This combines broad exploration of the configuration space with targeted sampling of regions that may be underrepresented by uniform sampling alone.

\subsubsection{Human lower-body motion}
Human reach is evaluated using the canonical human model~\cite{human_description} across a family of symmetric, flat-footed postures ranging from standing to squatting. At each pelvis height, we solve for a consistent leg configuration that respects joint limits and keeps both soles on the ground. Torso and arm configurations are then sampled conditional on these lower-body postures. This captures the contribution of lower-body motion to vertical reach while preserving a consistent relationship between pelvis height and leg articulation. Subsequent refinement remains on the same family of squat poses.

\subsubsection{Workspace discretization and refinement}
Endpoint positions are discretized on an axis-aligned grid of $25\,\mathrm{mm}$ cubic voxels. A voxel is occupied if it contains at least one sampled endpoint position, and its volume contributes to the corresponding arm's workspace. Left and right occupancy are maintained separately so that both combined reach and arm-specific correspondence can be evaluated.

We improve boundary coverage through adaptive local sampling. Occupied voxels adjacent to empty space identify candidate boundary configurations. We perturb the associated shared and arm joints within their limits, evaluate the new endpoint positions, and merge them into the occupancy sets. Progressively smaller perturbations refine the sampled boundary; refinement ends when its sampling budget is exhausted or too few new occupied voxels are discovered. All quantitative comparisons use the unsmoothed occupancy sets. Surface smoothing is used only to visualize workspace shape.

The reported comparison isolates positional kinematic reach subject to joint limits. It applies neither collision rejection nor static-stability filtering, and sampled endpoint positions below the ground plane are retained. These assumptions are shared across the workspace metrics.

\subsubsection{Workspace comparison metrics}
We compare the human and robot occupancy sets on a common $25\,\mathrm{mm}$ lattice. Let $H_L,H_R$ and $R_L,R_R$ denote their occupied voxel-center sets, with unions $H_U=H_L\cup H_R$ and $R_U=R_L\cup R_R$. For equal-sized voxels, ratios of cardinalities equal the corresponding volume ratios. Define the directed tolerance match
\begin{equation}
  \begin{aligned}
    d(\mathbf a,B)&=\min_{\mathbf b\in B}\|\mathbf a-\mathbf b\|_2,\\
    N_\tau(A,B)&=\bigl|\{\mathbf a\in A:d(\mathbf a,B)\leq\tau\}\bigr|,
    \qquad \tau=0.05\,\mathrm m.
  \end{aligned}
  \label{eq:workspace_tolerance}
\end{equation}
For the combined workspaces, the two directional match fractions are
\begin{equation}
  C_\tau^U=\frac{N_\tau(H_U,R_U)}{|H_U|},
  \qquad
  P_\tau^U=\frac{N_\tau(R_U,H_U)}{|R_U|}.
  \label{eq:workspace_union_match}
\end{equation}
Their harmonic mean gives Overall Reach Match. To preserve arm identity, we instead pool matched and total cell counts over corresponding arms:
\begin{equation}
  \begin{aligned}
    C_\tau^A&=\frac{\sum_{a\in\{L,R\}}N_\tau(H_a,R_a)}
    {|H_L|+|H_R|},\\
    P_\tau^A&=\frac{\sum_{a\in\{L,R\}}N_\tau(R_a,H_a)}
    {|R_L|+|R_R|}.
  \end{aligned}
  \label{eq:workspace_arm_match}
\end{equation}
The reported match scores are therefore
\begin{equation}
  M_{\rm overall}=\frac{2C_\tau^U P_\tau^U}{C_\tau^U+P_\tau^U},
  \qquad
  M_{\rm arm}=\frac{2C_\tau^A P_\tau^A}{C_\tau^A+P_\tau^A}.
  \label{eq:workspace_match_scores}
\end{equation}
These symmetric proximity scores penalize both missing human reach and robot reach outside the human workspace, allowing a $50\,\mathrm{mm}$ positional tolerance. Arm-specific pooling preserves the identity of the reaching arm and weights each side by its occupied-cell count. A zero denominator gives a score of zero.

Human Reach Coverage and Relative Workspace Volume retain exact voxel occupancy, without the tolerance:
\begin{equation}
  C_{\rm human}=\frac{|H_U\cap R_U|}{|H_U|},
  \qquad
  \rho_{\rm volume}=\frac{|R_U|}{|H_U|}.
  \label{eq:workspace_coverage_volume}
\end{equation}
These distinguish coverage of human reach from overall workspace size. The union and arm-specific sets describe marginal positional reach: two points, or a cell reachable by both arms, may require different body configurations. They do not establish simultaneous bimanual feasibility, end-effector orientation capability, or collision-free trajectories. Finite sampling and voxel resolution also limit the precision of the estimated boundaries.
\par
\endgroup

\begingroup
\color{black}
\subsection{Results}
\label{sec:appendix_results}

Tables~\ref{tab:appendix_results_dofs}--\ref{tab:appendix_results_times} provide the numerical results underlying the hardware, policy-design, and data-collection-interface evaluations in Figures~\ref{fig:hardware_test_results}, \ref{fig:policy_design_results}, and~\ref{fig:interface_design_results}. Task progress is expressed as a percentage using the scoring rubrics in Appendix~\ref{sec:appendix_task_scoring}.

For task-progress results, entries are mean $\pm$ sample standard deviation across trials; Average uses the mean and sample standard deviation across task means. Hardware ablations use five trials per task and condition, evaluated separately from the policy ablations. Human-data policies, robot-data policies, and co-training use 20 rollouts per task and condition. Observation ablations retain relative-base actions, while action ablations use all three cameras.

Gripper forces are in newtons. Pull-out tests use eight objects with three trials per object and direction, capped at $100\,\mathrm{N}$. Entries report the average object mean $\pm$ the average within-object sample standard deviation; closing force reports the mean $\pm$ sample standard deviation of three trials.

Completion times are in seconds and report mean $\pm$ sample standard deviation across all attempts. Unsuccessful attempts are assigned the longest successful time for the same interface and task.

\begin{table*}[!t]
  \centering
  \color{black}
  \caption{Hardware degree of freedom ablations.}
  \label{tab:appendix_results_dofs}
  \small
  \setlength{\tabcolsep}{6pt}
  \renewcommand{\arraystretch}{1.12}
  \begin{tabular*}{\textwidth}{@{\extracolsep{\fill}}lcccc@{}}
    \toprule
    \textbf{Configuration} & \textbf{Toy Cleanup} & \textbf{Tidy Pillow} & \textbf{Trash Disposal} & \textbf{Average} \\
    \midrule
    Complete system & $\boldsymbol{100.0 \pm 0.0}$ & $\boldsymbol{100.0 \pm 0.0}$ & $\boldsymbol{100.0 \pm 0.0}$ & $\boldsymbol{100.0 \pm 0.0}$ \\
    Torso disabled & $60.0 \pm 54.8$ & $14.3 \pm 0.0$ & $68.0 \pm 46.0$ & $47.4 \pm 29.0$ \\
    Lift disabled & $60.0 \pm 54.8$ & $85.7 \pm 24.7$ & $68.0 \pm 46.0$ & $71.2 \pm 13.2$ \\
    Head control disabled & $\boldsymbol{100.0 \pm 0.0}$ & $20.0 \pm 12.8$ & $36.0 \pm 41.0$ & $52.0 \pm 42.3$ \\
    \bottomrule
  \end{tabular*}
\end{table*}

\begin{table*}[!t]
  \centering
  \color{black}
  \caption{Policy ablations with sensorized human data.}
  \label{tab:appendix_results_human}
  \small
  \setlength{\tabcolsep}{6pt}
  \renewcommand{\arraystretch}{1.12}
  \begin{tabular*}{\textwidth}{@{\extracolsep{\fill}}lcccc@{}}
    \toprule
    \textbf{Configuration} & \textbf{Toy Cleanup} & \textbf{Tidy Pillow} & \textbf{Trash Disposal} & \textbf{Average} \\
    \midrule
    All cameras, relative-base & $75.0 \pm 44.4$ & $\boldsymbol{95.0 \pm 10.6}$ & $\boldsymbol{76.0 \pm 37.6}$ & $\boldsymbol{82.0 \pm 11.3}$ \\
    Head camera only & $5.0 \pm 15.4$ & $80.7 \pm 25.9$ & $34.0 \pm 42.6$ & $39.9 \pm 38.2$ \\
    Wrist cameras only & $82.5 \pm 37.3$ & $72.9 \pm 29.3$ & $44.0 \pm 41.3$ & $66.5 \pm 20.0$ \\
    Relative-pure actions & $\boldsymbol{85.0 \pm 36.6}$ & $47.9 \pm 22.4$ & $69.0 \pm 25.5$ & $67.3 \pm 18.6$ \\
    Relative-head actions & $20.0 \pm 41.0$ & $25.0 \pm 27.4$ & $32.0 \pm 37.5$ & $25.7 \pm 6.0$ \\
    \bottomrule
  \end{tabular*}
\end{table*}

\begin{table*}[!t]
  \centering
  \color{black}
  \caption{Policy ablations with robot teleoperation data.}
  \label{tab:appendix_results_robot}
  \small
  \setlength{\tabcolsep}{6pt}
  \renewcommand{\arraystretch}{1.12}
  \begin{tabular*}{\textwidth}{@{\extracolsep{\fill}}lcccc@{}}
    \toprule
    \textbf{Configuration} & \textbf{Vacuum Cleaning} & \textbf{Snack Retrieval} & \textbf{Hang Jacket} & \textbf{Average} \\
    \midrule
    All cameras, relative-base & $65.0 \pm 45.9$ & $88.3 \pm 27.3$ & $82.9 \pm 22.1$ & $78.7 \pm 12.2$ \\
    Head camera only & $0.0 \pm 0.0$ & $\boldsymbol{100.0 \pm 0.0}$ & $78.6 \pm 22.6$ & $59.5 \pm 52.7$ \\
    Wrist cameras only & $70.0 \pm 38.7$ & $93.3 \pm 14.1$ & $58.6 \pm 22.8$ & $74.0 \pm 17.7$ \\
    Absolute joint-space actions & $\boldsymbol{77.5 \pm 41.6}$ & $90.0 \pm 31.6$ & $\boldsymbol{91.4 \pm 18.1}$ & $\boldsymbol{86.3 \pm 7.7}$ \\
    \bottomrule
  \end{tabular*}
\end{table*}

\begin{table*}[!t]
  \centering
  \color{black}
  \caption{Human--robot co-training results.}
  \label{tab:appendix_results_cotraining}
  \small
  \setlength{\tabcolsep}{6pt}
  \renewcommand{\arraystretch}{1.12}
  \begin{tabular*}{\textwidth}{@{\extracolsep{\fill}}lccc@{}}
    \toprule
    \textbf{Training data} & \textbf{Toy Cleanup} & \textbf{Tidy Pillow} & \textbf{Average} \\
    \midrule
    Sensorized human only & $75.0 \pm 44.4$ & $\boldsymbol{95.0 \pm 10.6}$ & $85.0 \pm 14.1$ \\
    Human + robot (balanced sampling) & $\boldsymbol{95.0 \pm 22.4}$ & $93.6 \pm 21.0$ & $\boldsymbol{94.3 \pm 1.0}$ \\
    \bottomrule
  \end{tabular*}
\end{table*}

\begin{table*}[!t]
  \centering
  \color{black}
  \caption{Gripper force measurements.}
  \label{tab:appendix_results_gripper}
  \small
  \setlength{\tabcolsep}{6pt}
  \renewcommand{\arraystretch}{1.12}
  \begin{tabular*}{\textwidth}{@{\extracolsep{\fill}}lccc@{}}
    \toprule
    \textbf{Gripper} & \textbf{Horizontal pull-out} & \textbf{Vertical pull-out} & \textbf{Closing force} \\
    \midrule
    \robotname{} & $\boldsymbol{82.7 \pm 2.6}$ & $\boldsymbol{71.0 \pm 3.0}$ & $\boldsymbol{92.9 \pm 2.4}$ \\
    AgileX Piper & $46.4 \pm 3.4$ & $43.4 \pm 6.1$ & $63.3 \pm 1.5$ \\
    \bottomrule
  \end{tabular*}
\end{table*}

\begin{table*}[!t]
  \centering
  \color{black}
  \caption{User-study completion times.}
  \label{tab:appendix_results_times}
  \small
  \setlength{\tabcolsep}{6pt}
  \renewcommand{\arraystretch}{1.12}
  \begin{tabular*}{\textwidth}{@{\extracolsep{\fill}}lcc@{}}
    \toprule
    \textbf{Interface} & \textbf{Tidy Pillow} & \textbf{Wine-Glass Pouring} \\
    \midrule
    Sensorized Human & $\boldsymbol{8.4 \pm 1.7}$ & $\boldsymbol{6.7 \pm 1.9}$ \\
    VR Joystick & $39.0 \pm 14.9$ & $29.1 \pm 9.9$ \\
    Leader Arm & $35.3 \pm 8.5$ & $36.9 \pm 20.8$ \\
    VR Body Tracking & $52.1 \pm 18.7$ & $34.0 \pm 11.7$ \\
    \bottomrule
  \end{tabular*}
\end{table*}

\par
\endgroup

\end{document}